\documentclass[10pt,journal,compsoc]{IEEEtran}
\usepackage{amsmath,amsfonts}
\usepackage{algorithmic}
\usepackage{algorithm}
\usepackage{array}
\usepackage[caption=false,font=normalsize,labelfont=sf,textfont=sf]{subfig}
\usepackage{textcomp}
\usepackage{stfloats}
\usepackage{url}
\usepackage{verbatim}
\usepackage{graphicx}
\usepackage{multirow}
\usepackage{makecell}
\usepackage{colortbl}
\usepackage{xcolor}
\usepackage{cite}
\usepackage{xspace}
\usepackage{arydshln} 
\usepackage{booktabs}
\usepackage[numbers]{natbib}

\usepackage{amssymb}
\usepackage{pifont}
\usepackage{color}
\usepackage{newfloat}
\usepackage{listings}
\usepackage{times}
\usepackage{latexsym}
\usepackage{enumitem}
\usepackage{ragged2e}

\newcommand{\zqh}{\color{black}}
\newcommand{\revision}{\color{black}}

\newcommand{\roberta}{\textsc{RoBERTa}\xspace}
\newcommand{\bartre}{\textsc{BART\_Re}\xspace}
\newcommand{\bartcont}{\textsc{BART\_Cont}\xspace}
\newcommand{\bartde}{\textsc{BART\_DeNo}\xspace}
\newcommand{\bartjb}{\textsc{BART\_JoBa}\xspace}
\newcommand{\bartjp}{\textsc{BART\_JoPr}\xspace}
\newcommand{\tre}{\textsc{T5\_Re}\xspace}
\newcommand{\tcont}{\textsc{T5\_Cont}\xspace}
\newcommand{\tde}{\textsc{T5\_DeNo}\xspace}
\newcommand{\tjb}{\textsc{T5\_JoBa}\xspace}
\newcommand{\tjp}{\textsc{T5\_JoPr}\xspace}
\newcommand{\es}{\textsc{E2S2}\xspace}
\newcommand{\ie}{i.e.,\xspace}

\newcommand{\pchat}{\textsc{PChat}\xspace}
\newcommand{\daily}{\textsc{DaDi}\xspace}
\newcommand{\echat}{\textsc{EChat}\xspace}
\newcommand{\sig}{$^{\ddagger}$}

\begin{document}

\title{E2S2: Encoding-Enhanced Sequence-to-Sequence Pretraining for Language Understanding and Generation}

\author{
        Qihuang~Zhong,~\IEEEmembership{Member,~IEEE,}
        Liang~Ding,~\IEEEmembership{Member,~IEEE,}
        Juhua~Liu,~\IEEEmembership{Member,~IEEE,}
        Bo~Du,~\IEEEmembership{Senior~Member,~IEEE,}
        and~Dacheng~Tao,~\IEEEmembership{Fellow,~IEEE}% <-this % stops a space

\thanks{This work was supported in part by the National Key Research and Development Program of China under Grant 2023YFC2705700, and in part by the National Natural Science Foundation of China under Grant U23B2048, 62076186 and 62225113. The numerical calculations in this paper have been done on the supercomputing system in the Supercomputing Center of Wuhan University. Our source code and final models are publicly available at https://github.com/WHU-ZQH/E2S2. \textit{Corresponding Authors: Juhua Liu, Bo Du (e-mail: \{liujuhua, dubo\}@whu.edu.cn).}}

\thanks{Q. Zhong, J. Liu and B. Du are with the School of Computer Science, National Engineering Research Center for Multimedia Software, Institute of Artificial Intelligence, and Hubei Key Laboratory of Multimedia and Network Communication Engineering, Wuhan University, Wuhan, China (e-mail: \{zhongqihuang, liujuhua, dubo\}@whu.edu.cn).}

\thanks{ L. Ding and D. Tao are with the School of Computer Science, Faculty of Engineering, The University of Sydney, Australia (e-mail: liangding.liam@gmail.com; dacheng.tao@gmail.com).}
}

%\markboth{Journal of \LaTeX\ Class Files,~Vol.~14, No.~8, December~2022}%
%{Zhong \MakeLowercase{\textit{et al.}}: E2S2}


\IEEEtitleabstractindextext{%
\begin{abstract}
\justifying{
Sequence-to-sequence (seq2seq) learning is a popular fashion for large-scale pretraining language models. However, the previous seq2seq pretraining models generally focus on reconstructive objectives on the decoder side and neglect the effect of encoder-side supervision, which we argue may lead to sub-optimal performance.
To verify our hypothesis, we first empirically study the functionalities of the encoder and decoder in seq2seq pretrained language models, and find that the encoder takes an important but under-exploitation role than the decoder regarding the downstream performance and neuron activation. Therefore, we propose an encoding-enhanced seq2seq pretraining strategy, namely \es, which improves the seq2seq models via integrating more efficient self-supervised information into the encoders. Specifically, E2S2 adopts two self-supervised objectives on the encoder side from two aspects: 1) locally denoising the corrupted sentence (denoising objective); and 2) globally learning better sentence representations (contrastive objective). With the help of both objectives, the encoder can effectively distinguish the noise tokens and capture high-level (\textit{i.e.}, syntactic and semantic) knowledge, thus strengthening the  ability of seq2seq model to accurately achieve the conditional generation. On a large diversity of downstream natural language understanding and generation tasks, E2S2 dominantly improves the performance of its powerful backbone models, \textit{e.g.}, BART and T5.
For example, upon BART backbone, we achieve +1.1\% averaged gain on the general language understanding evaluation (GLUE) benchmark and +1.75\% $F_{0.5}$ score improvement on CoNLL2014 dataset. We also provide in-depth analyses to show the improvement stems from better linguistic representation. We hope that our work will foster future self-supervision research on seq2seq language model pretraining.}
\end{abstract}

}

\maketitle
\IEEEdisplaynontitleabstractindextext
\IEEEpeerreviewmaketitle

\IEEEraisesectionheading{\section{Introduction}\label{intro}}
\IEEEPARstart{S}{}equence-to-sequence (seq2seq) pretrained language models (PLMs)~\cite{ramachandran2017unsupervised,song2019mass,lewis2020bart,raffel2020exploring,qi2020prophetnet} are widely used in the community of natural language processing and have achieved remarkable success in numerous downstream tasks of both natural language generation (NLG) and understanding (NLU), such as machine translation \cite{liu2020multilingual,song2019mass,lewis2020pre}, text summarization \cite{wang2019denoising,qi2020prophetnet}, grammatical error correction \cite{zhou2021improving} and other discriminative tasks \cite{lewis2020bart,li2020survey,li2020neural}. Specifically, seq2seq models are generally implemented with an encoder-decoder framework \cite{sutskever2014sequence}, where the encoder models the input sentence first and then the decoder generates the output tokens auto-regressively conditioned on the representation of encoder.

\begin{figure}[t]
	\includegraphics[width=0.48\textwidth]{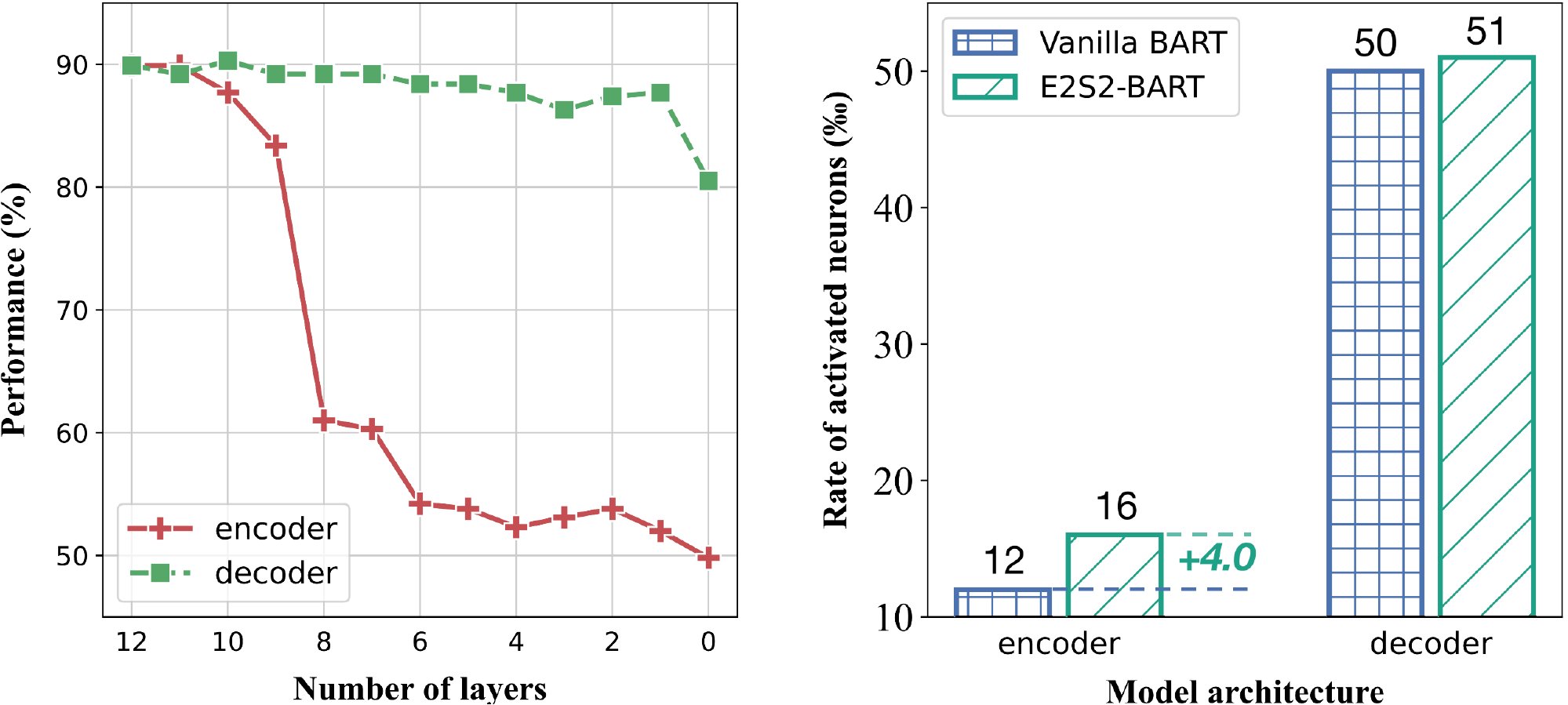} 
    \centering
	\caption{\textbf{Left}: comparison of downstream performance decrease when removing encoder and decoder layers of seq2seq PLM (\textit{i.e.}, BART~\cite{lewis2020bart}) respectively. \textbf{Right}: the average rate of activated neurons of FFN layers in the encoder and decoder, where the higher rate denotes the FFN trained more sufficiently. \textbf{Takeaway:} \textit{the encoder plays a key role in the seq2seq framework, but its training is less sufficient than that of decoder.}}
	\label{e2s2:fig1}
\end{figure}

The large-scale seq2seq PLMs are basically trained with reconstructive
% denoising auto-encoding 
self-supervisions. Concretely, the encoder receives the perturbed input with different token- and sentence-level noising functions~\cite{song2019mass,raffel2020exploring,lewis2020bart}, \textit{e.g.}, text infilling, token deletion, token masking, and sentence permutation. Then the decoder reconstructs the perturbed text. Such a self-supervised learning process is assumed to learn the knowledge contained in the large-scale text, thus providing better initialization for downstream tasks.
For instance, MASS~\cite{song2019mass} takes the sentence with randomly masked fragment as input of encoder and makes the decoder predict this masked fragment. BART~\cite{lewis2020bart} improves the MASS via predicting the complete original sentence, instead of the missing tokens. 

Although the remarkable progress of seq2seq pretraining has been witnessed, we find that most current seq2seq PLMs usually achieve suboptimal performance on downstream language understanding tasks. An obvious evidence is that the top-ranked systems of the General Language Understanding Evaluation (GLUE) leaderboard\footnote{GLUE is a popular NLU benchmark that is used to evaluate the performance of PLMs (see section~\ref{dataset}), and the leaderboard can be seen in \url{https://gluebenchmark.com/leaderboard} .} are almost (encoder-only) masked language models, \textit{i.e.}, BERT \cite{devlin2019bert} and its variants~\cite{liu2019roberta,he2020deberta}. 
We attribute this phenomenon to the unsatisfactory capability of encoder in seq2seq PLMs, as most of them are trained via optimizing a reconstruction loss on the decoder side, but neglecting the impact of self-supervised information on the encoder side. 
To verify our hypothesis, we empirically study the functionalities of the encoder and decoder in the seq2seq PLM (\textit{i.e.}, BART~\cite{lewis2020bart}), as shown the results in Fig.~\ref{e2s2:fig1}. 
Specifically, in Fig.~\ref{e2s2:fig1} (\textbf{Left}), {\zqh we remove the encoder and decoder layers of pretrained BART, respectively, and analyze the performance decrease. As seen,} there is only a slight performance decrease when removing decoder layers. However, if we remove several encoder layers, the performance drops dramatically, which indicates that encoders have a greater influence on seq2seq models, {\zqh which is similar to the findings of Kasai~et~al.~\cite{kasai2021deep} that confirm using deep encoders and shallow decoders is more effective in seq2seq scenarios}. Additionally, results in Fig.~\ref{e2s2:fig1} (\textbf{Right}) show that the encoder of seq2seq PLM is under-exploited, as the rate of activated neurons of encoder is much lower than those of decoder.
In general, from these results, we can basically conclude that \textit{\textbf{
the encoder takes an important but under-exploitation role than the decoder in seq2seq PLMs
}}.

To this end, we propose an encoding-enhanced seq2seq pertaining strategy, denoted as \textit{\textbf{E2S2}}, to provide richer supervision for the seq2seq models via integrating two self-supervised objectives on the side of encoder. 
Specifically, we introduce the self-supervised information from two aspects: 
1) \textbf{\textit{locally}} denoising the corrupted sentence earlier (\textit{denoising objective}); 
2) \textbf{\textit{globally}} learning better sentence representations (\textit{contrastive objective}). 
Firstly, given the input sentences corrupted with various noising functions, different from the vanilla seq2seq PLMs that only reconstruct the correct sentences on the decoder side, our motivation is to encourage the encoder to effectively distinguish the corrupted tokens earlier and then help the decoder better perform the reconstruction process. 
On the other hand, inspired by Li~et~al.~\cite{li2020sentence} that find the encoder always induces a non-smooth anisotropic semantic space of sentences, which harms the performance of sentence representation, we further present a  contrastive objective to globally improve the sentence representations learned by the encoder. In practice, for contrastive objective, we provide two solutions to obtain the sentence representations. The first is simply average pooling on all hidden representations of the last encoder layer. The second is to employ a more effective prompt-based sentence construction approach to acquire more reliable representations. In this way, the encoder is forced to learn better sentence representation, from which the decoder can acquire more knowledge and perform better on language understanding and generation.

Extensive experiments on several downstream tasks, including both NLU (\textit{i.e.}, GLUE benchmark~\cite{wang2018glue}) and NLG (text summarization, grammatical error correction, and dialog generation), show that E2S2 consistently outperforms the vanilla seq2seq pretraining scheme. More specifically, using the typical seq2seq PLM (BART~\cite{lewis2020bart}) as the baseline model, E2S2 achieves +1.1\% averaged performance improvement on test sets of GLUE benchmark, especially +2.3\% improvement on the CoLA task that is highly related to the denoising objective. In addition to the positive effect on NLU tasks, we also observe consistent improvements on NLG tasks, \textit{e.g.}, +1.75\% $F_{0.5}$ score for grammatical error correction task. We also empirically prove that our E2S2 strategy is compatible with other seq2seq PLMs, \textit{e.g.}, T5~\cite{raffel2020exploring}.
These results demonstrate the effectiveness and universality of our E2S2 strategy. Additionally, in-depth discussions in section~\ref{disscusion} prove that E2S2 indeed enables the encoder to learn better linguistical representations.

Our main contributions can be summarized as follows:
\begin{itemize}
    \item We explore the shortcomings (unsatisfactory capability of the encoder) of seq2seq pretraining and present an encoder-enhanced learning strategy (termed as E2S2) to recast the vanilla seq2seq pretraining scheme via integrating more self-supervised information on the side of encoder.

    \item We design two self-supervised pretraining objectives from two meaningful aspects, \textit{i.e.}, locally denoising the perturbed text earlier and globally learning better sentence representations, which are easy-to-implement and model-agnostic.
    
    \item Extensive experiments upon both NLU and NLG tasks validate the effectiveness and universality of E2S2. Quantitative analysis and in-depth discussion provide some insights into where the improvements come from.
\end{itemize}

The rest of this paper is organized as follows. In section~\ref{sec:related work}, we briefly review the related works. In section~\ref{sec:method}, we introduce our proposed method in detail. Section~\ref{sec:experiment} reports and discusses our experimental results. Finally, we conclude our study in section~\ref{sec:conclusion}.

\section{Related Works}
\label{sec:related work}
\subsection{Pretrained Language Models}
In recent years, we have witnessed numerous pretrained language models (PLMs) that achieved tremendous success in the community of NLP. Based on the model architectures, these PLMs can be classified into three main categories: decoder-only (auto-regressive) LMs, encoder-only (auto-encoding) LMs and encoder-decoder (sequence-to-sequence) LMs. Auto-regressive LMs aim to predict the future words towards a sequence of words, such as GPT~\cite{radford2018improving} and its variants. Such auto-regressive models are well-suitable for language generation, but they are unidirectional and usually fail short in the representation learning for understanding the sentence~\cite{liu2021pre}. Thus, researchers turn to focus on auto-encoding LMs that introduce a bidirectional masked language modeling (MLM) objective to predict the masked text token based on the context. The most representative auto-encoding LMs are BERT~\cite{devlin2019bert} and its variants, \textit{e.g.}, \roberta~\cite{liu2019roberta} and DeBERTa~\cite{he2020deberta}. 

In order to combine the advantages of auto-regressive LMs and auto-encoding LMs, seq2seq LMs are sequentially proposed, which firstly employ a separate encoder to model the source text and then use a left-to-right LM to decode the conditional target text. For example, as the typical seq2seq LMs, BART \cite{lewis2020bart} and T5 \cite{raffel2020exploring} first corrupt the text with various noising functions on the encoder side and then train the models to reconstruct the original text in an auto-regressive manner. The encoder-decoder paradigm makes the seq2seq LMs not only generally suitable for text generation, but also well for text understanding tasks. In practice, for text understanding tasks, BART employs the final hidden state of the final decoder token as the sentence representation and introduces an additional MLP layer to output the prediction, while T5 converts the classification tasks as ``text-to-text" generation tasks and directly generates the target texts, \textit{e.g.}, sentiment polarity for sentiment analysis task.

The above seq2seq PLMs have achieved remarkable progress in various NLP tasks, but there are still some limitations that need to be improved. In particular, when seq2seq PLMs are used to process language understanding tasks, their performance is usually worse than those of auto-encoding LMs. One possible reason is that the encoder of seq2seq LMs is not trained sufficiently and the representations given by the encoder are suboptimal. Intuitively, enhancing the encoder is beneficial for the pretraining of seq2seq LMs. To the best of our knowledge, our E2S2 is one of rare works that propose to integrate more self-supervised information into the encoder optimization in the encoder-decoder paradigm.

\subsection{Self-supervision Learning in PLMs} Self-supervised learning (SSL) is the de facto standard for large-scale pretraining in the NLP community~\cite{wu2021self,liu2021self,liu2022graph,zhong2023self,zhong2023revisiting}, which helps model to learn universal knowledge based on the (pseudo) self-supervision provided by pretraining tasks. In general, self-supervised learning for PLMs can be classified into generative SSL, contrastive SSL and adversarial SSL. The generative SSL aims to train the models by decoding the encoded inputs. The most representative models are GPT~\cite{radford2018improving} and BERT~\cite{devlin2019bert}. The former predicts the next tokens based on the previous tokens, while the later predicts random masked tokens based on the (bidirectional) unmasked tokens (token-level MLM). Motivated by GPT and BERT, more PLMs involving generative SSL are further proposed, \textit{e.g.}, SpanBERT~\cite{joshi2020spanbert} (span-level MLM), ERNIE~\cite{sun2020ernie} (entity-level MLM) and XLNet~\cite{yang2019xlnet} (permutation language modeling).

Contrastive SSL refers to training the models by comparing, such as the next sentence prediction (NSP) objective in BERT that aims to distinguish whether the given sentence pair is consecutive. In addition to such a \texttt{context-instance} contrast, a breakthrough in contrastive SSL for PLMs has recently been achieved by \texttt{instance-instance} contrastive learning, \textit{e.g.}, SimCSE~\cite{gao2021simcse}, ConSERT~\cite{yan2021consert} and PromptBERT~\cite{jiang2022promptbert}, which encourages the PLMs to ``learn to compare'' by a noise-contrastive estimation. For the adversarial SSL, the models are enforced to identify whether the input tokens are replaced or shuffled. The typical adversarial SSL includes the relaced token detection~\cite{clark2019electra}, shuffled token detection~\cite{panda2021shuffled}, etc.

As aforementioned above, our work aims to integrate more self-supervised information into the encoder training. Specifically, in addition to the generative SSL (text infilling) on the decoder side, our approach involves enriching the semantic and linguistical knowledge of encoder by adding extra SSL supervision in a \textit{local-to-global} manner, \textit{i.e.}, locally detecting the shuffle/random token (\textit{i.e.}, denoising objective) and globally learning better sentence representations (\textit{i.e.}, contrastive objective).

\begin{figure*}[t]
	\includegraphics[width=1\textwidth]{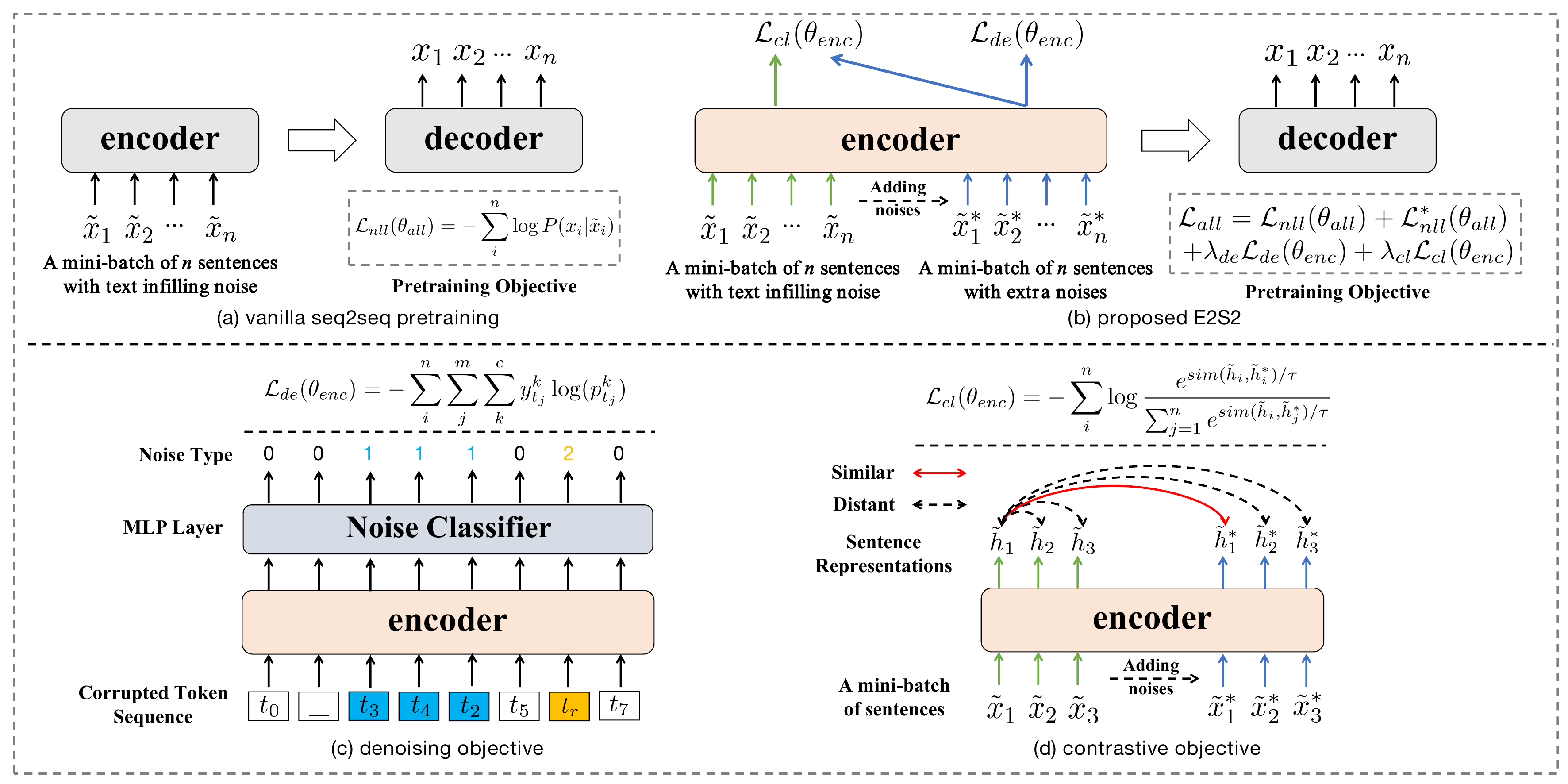} 
	\caption{The schematic comparison of our E2S2 with the vanilla seq2seq pretraining scheme. In general, \textbf{(a)} is the original seq2seq pretraining scheme and \textbf{(b)} is our proposed E2S2 strategy, which integrates two self-supervised objectives on the encoder side, \textit{i.e.}, denoising objective $\mathcal{L}_{de}$ and contrastive objective $\mathcal{L}_{cl}$. Notably, $\tilde{x}$ is the sentence corrupted with text infilling noise, $\tilde{x}^*$ is corrupted with extra noises (\textit{e.g.}, shuffling and random replacement), $n$ is the number of sentences in a mini-batch, $\theta_{enc}$ and $\theta_{all}$ denote the parameters of the encoder and full model respectively. 
    Specifically, \textbf{(c)} and \textbf{(d)} provide a more detailed illustration for each objective. In \textbf{(c)}, the token sequence is obtained by corrupting the original input tokens $\{t_0,t_1,t_2,t_3,t_4,t_5,t_6,t_7\}$, where the boxes in blue denote the shuffled tokens and the yellow box denotes the randomly replaced token. $m$ and $c$ are the length of token sequence and the classes of noise type (1 for shuffle, 2 for random replacement, and 0 for others) respectively. $y$ and $p$ are the ground-truths and predictions of the noise type.}
	\label{e2s2:fig2}
\end{figure*}

\section{Methodology}
\label{sec:method}
In this section, we present the introduction of background, and then describe our proposed E2S2 strategy in detail.

\subsection{Background}
\subsubsection{Sequence-to-Sequence Pretraining}
Suppose we have a source sentence $x$, the goal of seq2seq pretraining is to train a seq2seq model by corrupting sentences and then optimizing the reconstruction loss. Specifically, we first apply noising schemes (\textit{e.g.}, token masking, text infilling, sentence permutation, etc.) to $x$ for obtaining the perturbed sentence $\tilde{x}$, which is then fed into the encoder to obtain the hidden representations $\tilde{h}$, and finally input $\tilde{h}$ into the decoder to auto-regressively reconstruct the original sentence $x$. Therefore, we can optimize the seq2seq model via maximizing the training objective, which is generally the log-likelihood of the sequence pairs $\{\tilde{x},x\}$ and can be defined as follows:
\begin{equation} \label{eq1}
% \mathcal{L}_{nll}(\theta_{all})=argmax_\theta log P(y|x;\theta_{all})
\mathcal{L}_{nll}(\theta_{all}) = -\sum^n_i \log P(x_i|\tilde{x}_i;\theta_{all}),
\end{equation}
where the $\theta_{all}$ denotes the parameters of full model, $n$ is the number of training samples in a mini-batch and $P(\cdot)$ is the predicted probability.

\subsubsection{Contrastive Learning}
Contrastive learning aims to obtain discriminative representations via mapping similar input sentences to nearby points in the output representation space, while mapping dissimilar input sentences to distant points~\cite{gao2021simcse}. Mathematically, given a set of paired samples $\{(x_i,x^+_i)\}|^n_i$, where $(x_i,x^+_i)$ is a pair of semantically similar samples, we can obtain the corresponding hidden representations $\{(h_i,h^+_i)\}|^n_i$. Notably, we follow the works \cite{gao2021simcse,yan2021consert} and assume the samples from different pairs are negatives, \textit{i.e.}, $(x_i,x_j)$, $(x_i,x^+_j)$ and $(x^+_i,x^+_j)$ are negative pairs, where $i$ is not equal to $j$. For a mini-batch with $n$ pairs of samples, the training objective of contrastive learning can be formulated as follows: 
\begin{equation} \label{eq2}
\mathcal{L}_{cl}=-\sum^n_i \log \frac{e^{sim(h_i,h^+_i)/\tau}}{\sum^n_{j=1} e^{sim(h_i, h^+_j)/\tau}},
\end{equation}
where $sim(\cdot)$ denotes the cosine similarity function, $\tau$ is the temperature factor and empirically set to 1 in this paper. 

\subsubsection{Prompt-based Learning}
Prompt-based learning attempts to integrate extra information by prepending textual prompts before the inputs and help the model to generate desired outputs of NLP tasks directly~\cite{lester2021power}.
The critical problem in prompt-based learning is how to design the prompt template. In many previous works \cite{brown2020language,schick2020few,schick2021exploiting}, the prompt is created based on human introspection and is normally a hard template, which is a textual string that has two main slots: one is the input slot \texttt{[X]} for input $x$ and the other is the representation slot \texttt{[Z]} for $x$ representation. Taking the sentiment analysis task as an example, given a sentence $x$ of ``The food is delicious." and a prompt of ``\texttt{[X]} Overall, it is \texttt{[Z]}.", we first fill the slot \texttt{[X]} with the input sentence $x$ and then obtain its corresponding representations, \textit{i.e.}, the output of \texttt{[Z]}, which can be further used to predict the target polarity.  
Such a simple prompt-based approach is easier to obtain the representations that contain richer linguistical knowledge from PLMs, as proven by the previous study~\cite{jiang2022promptbert}.

\subsection{Encoding-Enhanced Sequence-to-Sequence Pretraining}
\label{e2s2}
As stated in section~\ref{intro}, we propose an encoding-enhanced strategy (E2S2) to improve the vanilla seq2seq pretrained models. To have a close look, we show a schematic comparison between the vanilla seq2seq pretraining scheme and our E2S2 scheme in Fig.~\ref{e2s2:fig2}. 
In particular, the major difference is the adding of several self-supervisions on the encoder side, which work from two aspects, \textit{i.e.}, locally denoising the perturbed sentences (denoising objective) and globally learning better sentence embeddings (contrastive objective). Taking the representative seq2seq model BART as our baseline, we sequentially introduce the processes of these two perspectives.

\subsubsection{Locally Denoising the Perturbed Sentences}
As mentioned above, the input sequence for training seq2seq PLMs is the perturbed sentence $\tilde{x}$, which is corrupted by noising functions. Notably, we adopt text infilling, 
{\zqh where a number of text spans are sampled and each span is replaced with a single \texttt{[MASK]} token,} 
as the basic nosing function\footnote{Although Lewis~et~al.~\cite{lewis2020bart} designed five noising functions in their original paper, for simplification we only keep the most effective one -- text infilling. {\zqh In our preliminary experiments, we found that continuing pretraining the BART with the simplified objective achieved the comparable downstream performance (88.71 \textit{vs.} 88.75 average score on GLUE benchmark) against the counterpart pretraining with full objectives.}}. {\zqh Text infilling teaches the model to predict how many tokens are
missing from a span.}
Besides, we introduce two additional linguistically motivated noising alternatives, \textit{i.e.}, \textit{Shuffle} and \textit{Random}\footnote{{\zqh Notably, there are more sophisticated noising functions, such as adversarial attack~\cite{morris2020textattack}. However, our main focus is to incorporate more supervised signals into the encoder, instead of exploring more complex noising functions. Thus, we simply use the widely-used \textit{Shuffle} and \textit{Random} in this paper.}}, which have proved to be beneficial to language models~\cite{yamaguchi2021frustratingly,alajrami2022does}. \textit{Shuffle} refers to selecting some trigrams from unmasked tokens and then shuffling these tokens in each trigram, and \textit{Random} refers to randomly replacing tokens with out-of-sequence tokens from the vocabulary. To distinguish different noising schemes, we denote the sequence corrupted by text infilling as $\tilde{x}$ and that corrupted by the extended noising scheme (\textit{i.e.}, the combination of text infilling, Shuffle and Random) as $\tilde{x}^*$.

Subsequently, we feed the corrupted sentence $\tilde{x}^*$ into the encoder and obtain the hidden representations $\tilde{h}^*$. Then, motivated by the success of ELECTRA~\cite{clark2019electra}, we employ a separate MLP classifier to detect whether each token of $x$ is shuffled or replaced, instead of directly predicting the original tokens. Such a process has been demonstrated more computationally cheaper and more simple-efficient in previous works \cite{nijkamp2021script,yamaguchi2021frustratingly}. The training objective is lastly built by minimizing the cross-entropy loss as formulated:
\begin{equation} \label{eq3}
\mathcal{L}_{de}(\theta_{enc})=-\sum^n_i \sum^m_j \sum^c_k \hat{y}^k_{t_j} \log (p^k_{t_j}),
\end{equation}
where $\hat{y}$ and $p$ denote the ground-truths and predictions of the noise type (\textit{i.e.}, 1 for shuffle, 2 for randomly replacement and 0 for others, as shown in Fig.~\ref{e2s2:fig2}(c)) . $m$ and $c$ are the length of the $\tilde{x}^*_i$ and the classes of noise type.

During training, the encoder is forced to acquire both syntactic and semantic knowledge by locally distinguishing between shuffled and replaced tokens in context. Therefore, the pretrained seq2seq model could effectively identify the noises from the input sentence, and perform better on downstream sequence rewriting tasks, \textit{e.g.}, grammatical error correction. 
This training objective performs in a denoising manner and enhances the encoder with more syntactic knowledge, thus we denote it as the denoising objective.

\subsubsection{Globally Learning Better Sentence Embeddings}
In addition to the objective of denoising the perturbed text, we also explore the poor performance of sentence embeddings on the encoder side. In particular, common wisdom in representation learning states that the contextualized language models, \textit{e.g.}, BERT, underperform in sentence embedding \cite{li2020sentence}, due to the issue of representation collapse, \textit{i.e.}, anisotropy. In sequence-to-sequence learning, such poor sentence representations learned by the encoder and subsequently fed into the decoder, would greatly hinder the performance of various NLU and NLG tasks.

To address the problem, we design a contrastive pretraining objective, which globally improves sentence embeddings with contrastive learning. 
There are two keys to implement our contrastive objective: 1) how to construct positive instances; and 2) how to obtain sentence embeddings from the output of encoder. Regarding the former, we follow Yan~et~al.~\cite{yan2021consert} and apply different noising functions on the input tokens to construct positive instances. In practice, given a sequence $x$, we simply feed it into the encoder twice, where one sequence is augmented with the introduced noising functions\footnote{{\zqh In some cases, shuffling or random replacing might change the original meaning of the sentence, which would cause the ``false'' positive pairs. Nevertheless, considering that it's complex and time-consuming to filter these special cases, we simply regard these cases as the (noisy) positive pairs and use them to boost the robustness of pretraining.}} (\textit{i.e.}, \textit{Shuffle} and \textit{Random}, obtaining $\tilde{x}^*$), and the other is augmented with the original text infilling (denoted as $\tilde{x}$), \textit{i.e.}, $\{\tilde{x}, \tilde{x}^*\}$ is a pair of positive instances. 

As for the latter, we provide two solutions to obtain sentence embeddings. The first is to simply use basic pooling operations, \textit{e.g.}, mean pooling, to process the hidden representations on the encoder side. The second is inspired by recent prompt-based studies, we introduce the prompt-based learning to acquire more reliable sentence representations. Specifically, we first create a manual prompt template, \textit{e.g.}, ``\texttt{[MASK]} is the representation of the sentence: \texttt{[X]}.", where \texttt{[X]} denotes the input slot. Then, we fill the slot \texttt{[X]} with the input sentence $\tilde{x}^*$ and feed the filled template into the encoder. {\zqh Note that the prompt templates are not included as the input in the decoder, \textit{i.e.}, the input of the decoder is still the original sentence.} Lastly, the hidden vector of \texttt{[MASK]} token on the encoder side is used to represent the sentence embeddings. Notably, while the prompt-based learning is proved more efficient than the first solution, the performance of prompt-based method is unstable and depends on the quality of the manual prompt template. The in-depth discussions on the impact of prompt-based methods can be referred to section~\ref{prompt}. 

In this way, we can construct the positive pairs $\{\tilde{x}, \tilde{x}^*\}$ and obtain the sentence representations of the encoder $\{(\tilde{h_i},\tilde{h}^*_i)\}$, respectively. Following Eq.~\ref{eq2}, we can achieve the introduced contrastive objective.

\subsubsection{\zqh Overall pretraining objective}
Combining all training objectives, the overall pretraining objective function of our proposed E2S2 can be formulated as follows:
\begin{equation} \label{eq4}
    \begin{split}
        \mathcal{L}_{all} &= \mathcal{L}^*_{nll}(\theta_{all})+\mathcal{L}_{nll}(\theta_{all})
        \\&+\lambda_{de} \mathcal{L}_{de}(\theta_{enc})+\lambda_{cl}\mathcal{L}_{cl}(\theta_{enc}),
    \end{split}
\end{equation}
where $\mathcal{L}^*_{nll}$ and $\mathcal{L}_{nll}$ are the vanilla reconstruct loss on the perturbed $\tilde{x}^*$ and $\tilde{x}$ respectively, and $\theta_{enc}$ denotes the parameters of encoder. $\lambda_{de}$ and $\lambda_{cl}$ are hyper-parameters used to balance the weights of different objectives, and set to 0.05 and 0.1, respectively. {\zqh The analysis of hyper-parameter can be seen in section~\ref{sec:parameter_analysis}.}

\begin{figure}[t]
\centering
\includegraphics[width=0.47\textwidth]{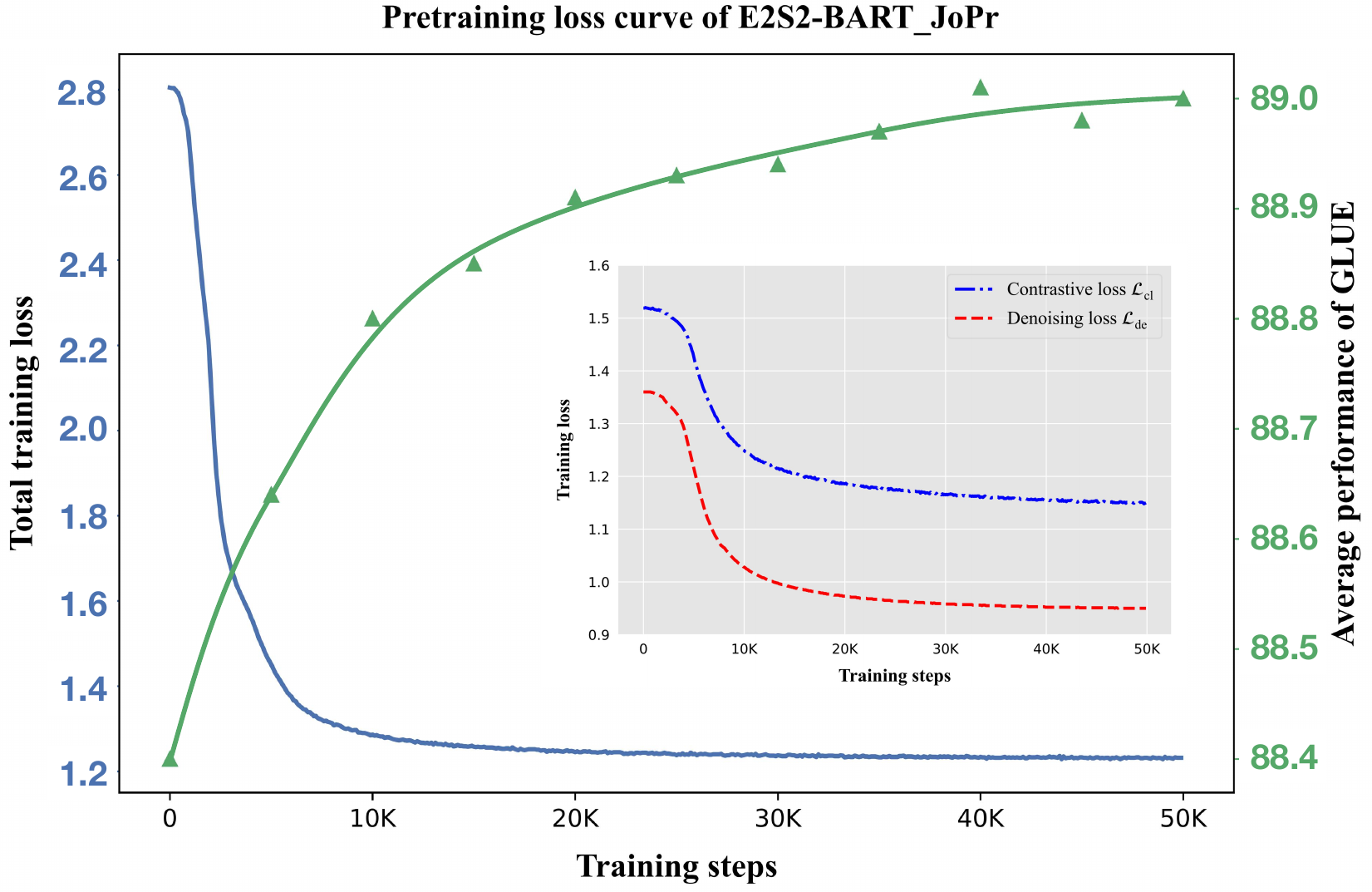}
\caption{\zqh Illustration of pretraining details. The left y-axis is the overall training loss, while the right y-axis is the average performance on the dev sets of GLUE. The loss curves of contrastive and denoising objectives are illustrated in the inserted figure.}
\label{fig:pretraining_loss}
\end{figure}

{\zqh To have a closer look, we illustrate the pretraining details, \textit{i.e.}, training loss curves and downstream performance across different pretraining steps, in Fig.~\ref{fig:pretraining_loss}. As seen, in the early stages of pretraining, the losses for each training objective were rapidly decreasing. Then, in the later stages of training, they gradually converged and reached stability in the final phase. Moreover, as training progresses, the performance of models on downstream tasks steadily improves, indicating the effectiveness of our E2S2 pretraining.
}

\begin{table}[t]
    \caption{Data divisions upon different downstream tasks.}
    \label{tab:dataset}
    \centering
    \begin{tabular}{cllll}
        \toprule
        \textbf{Task}                    & \textbf{Dataset}             & \textbf{\#Train}   & \textbf{\#Valid} & \textbf{\#Test}  \\
        \midrule \midrule
        \multirow{8}{*}{\textbf{GLUE}}   & CoLA                & 8,551     & 1,043   & 1,063   \\
                                & SST                 & 67,350    & 873     & 1,821   \\
                                & MRPC                & 5,801     & 4,076   & 1,725   \\
                                & STS-B               & 5,749     & 1,500   & 1,379   \\
                                & QQP                 & 363,870   & 40,431  & 390,956 \\
                                & MNLI                & 392,720   & 9,815   & 9,796   \\
                                & QNLI                & 104,743   & 5,463   & 5,461   \\
                                & RTE                 & 2,491     & 277     & 3,000   \\ \cmidrule{2-5}
        \multirow{3}{*}{\textbf{\textsc{AbsSum}}} & CNN/DM              & 287,227   & 13,368  & 11,490  \\
                                & XSum                & 204,045   & 11,332  & 11,334  \\
                                & SAMSum              & 14,732    & 818     & 819     \\ \cmidrule{2-5}
        \textbf{GEC}                     & CoNLL2014           & 1,298,756 & 5,448   & 1,312   \\ \cmidrule{2-5}
        \multirow{4}{*}{\textbf{\textsc{DiaGe}}}  & PersonalChat        & 122,499   & 14,602  & 14,056  \\ 
                                & DailyDialog         & 76,052    & 7,069   & 6,740   \\
                                & DSTC7               & 76,590    & 17,870  & 1,710   \\
                                & EmpatheticDialogues & 64,633    & 9,305   & 8,423  \\
        \bottomrule
        \end{tabular}
\end{table}
\begin{table*}[ht]
    \caption{Experimental settings on all downstream tasks. The upper are NLU tasks, while the bottom are NLG tasks. {\zqh Notably, for NLG tasks, we use the ``max\_tokens'' to express the batch size in number of tokens.}}
    \label{tab:setting}
    \centering
    \scalebox{1.2}{
    \begin{tabular}{lllllllll}
    \toprule
    \multicolumn{1}{c}{}                   & \textbf{MNLI}    & \textbf{QNLI}    & \textbf{QQP}     & \textbf{RTE}   & \textbf{SST-2}  & \textbf{MRPC}  & \textbf{CoLA}  & \textbf{STS-B}  \\ \midrule
    --num\_classes      & 3       & 2       & 2       & 2     & 2      & 2     & 2     & 1           \\
    --learning\_rate    & 5e-6    & 1e-5    & 1e-5    & 1e-5  & 5e-6   & 2e-5  & 2e-5  & 2e-5    \\
    --batch\_size       & 256     & 64      & 96      & 32    & 128    & 64    & 64    & 32     \\
    --total\_steps & 15,484   & 16,556   & 37,757   & 1,018  & 5,233   & 1,148  & 1,334  & 1,799       \\
    --warmup\_steps    & 929     & 993     & 2,265    & 61    & 314    & 68    & 80    & 107     \\
    --num\_GPUs         & 4       & 4       & 6       & 2     & 2      & 2     & 2     & 2       \\ \midrule \midrule
                       & \textbf{CNN/DM}    & \textbf{XSum}    & \textbf{SAMSum}     & \textbf{Conll14}   & \textbf{\pchat}  & \textbf{\daily}  & \textbf{DSTC7}  & \textbf{\echat}  \\ \midrule
    --learning\_rate    & 3e-5    & 3e-5    & 4e-5    & 5e-4  &4e-5 &5e-5  &4e-5  & 5e-5    \\
    --max\_tokens       & 8,192    & 4,096     &4,096       & 16,384    &4,096    &4,096    &4,096    &4,096    \\
    --total\_steps & 20,000   & 15,000   &1,000   & 2,000  & 1,000   &1,000   &1,000   &2,000    \\
    --warmup\_steps    & 500     & 500     &200     & 120    &200     &200     &200     &200      \\
    --num\_GPUs         & 8       & 8       &8        & 4     & 8      & 8     &8      &8     \\
    \bottomrule
    \end{tabular}
    }
\end{table*}

\section{Experiments}
\label{sec:experiment}
In this section, we conduct extensive experiments to investigate the general language learning abilities of our proposed E2S2 strategy. Specifically, we evaluate the models trained with E2S2 on a diverse of downstream benchmarks, covering representative tasks from the fields of both language understanding and generation.

\subsection{Tasks and Datasets}
\label{dataset}
\subsubsection{\textbf{Discriminative Tasks}} The discriminative tasks refer to making the corresponding predictions towards single-sentence or sentence-pair inputs. To provide a comprehensive comparison with the baselines, we report the results on the popular benchmark, \ie General Language Understanding Evaluation (GLUE)
~\cite{wang2018glue}, which consists of several language understanding tasks including sentiment analysis, question answering and textual entailment. The benchmark provides the resources of training and evaluating for each task, and a public leaderboard for analyzing the systems on the private test data. 
For fair comparison, we respectively report the results on the development and test sets of single-task, \ie without multi-task or ensemble training. In practice, we evaluate the performance with Accuracy (``\textit{Acc}'') metric for most tasks, except the F1 scores for QQP and MRPC, the Pearson-Spearman correlations (``\textit{Pcor/Scor}'') for STS-B and the Matthew correlations (``\textit{Mcc}'') for CoLA.

\subsubsection{\textbf{Abstractive Summarization}} Given a long document, the abstractive summarization (\textsc{AbsSum}) aims to convert it into a short and adequate summary in the same language. For this task, we employ three widely-used summarization datasets, \ie CNN/DM\footnote{\url{https://huggingface.co/datasets/cnn_dailymail}}, XSum and SAMSum\footnote{\url{https://github.com/Junpliu/ConDigSum}}. The CNN/DM~\cite{hermann2015teaching} and XSum~\cite{narayan2018don} contain 287K and 204K document-summary pairs respectively, while the SAMSum~\cite{liu2021topic} consists of 14K dialogue-summary pairs for training. We follow the prior studies~\cite{Ott:19, rothe2020leveraging,zhong2022improving} to preprocess the data. During testing, for the CNN/DM, the minimum and maximum lengths were respectively set to 55 and 140, which were tuned on the development data. Similarly, we empirically set the minimum/maximum length of XSum and SAMSum as 10/60 and 5/100, respectively. Notably, the target summaries are closely related to source documents in CNN/DM, while the target summaries in XSum are more abstractive. Following the recent works, we report evaluation results in terms of the standard ROUGE metric, \textit{i.e.}, Rouge-1, Rouge-2 and Rouge-L, respectively. 

\subsubsection{\textbf{Grammatical Error Correction}} The grammatical error correction (GEC) task aims to rewrite the input sentence with grammatical errors into the corresponding correct sentence, where the original and target sentences have the similar sentence lengths \cite{ng2014conll}. We conduct comparison experiments on the representative GEC benchmark, \textit{i.e.}, CoNLL2014\footnote{\url{https://www.comp.nus.edu.sg/~nlp/conll14st.html}}, which contains 1.4M, 5K and 1K training, validation and test samples, respectively. We closely follow Chollampatt~et~al.~\cite{Chollampatt:18} to preprocess the data. The MaxMatch ($\text{M}^2$) scores~\cite{Dahlmeier:12} are used for evaluation with Precision and $F_{0.5}$ values. 

\subsubsection{\textbf{Dialogue Generation}} The dialogue generation (\textsc{DiaGe}) task refers to generating meaningful and coherent dialogue responses based on the dialogue history. We use four popular \textsc{DiaGe} public datasets as benchmarks, covering PersonalChat~\cite{zhang2018personalizing} (denoted as \pchat), DailyDialog~\cite{lidailydialog} (denoted as \daily), DSTC7~\cite{alamri2019audio} and EmpatheticDialogues~\cite{rashkin2019towards} (denoted as \echat). For this task, 
% we simply evaluate the performance on the PPL 
{\revision we report the PPL of validation test and use the BLEU~\cite{papineni2002bleu} metric to evaluate the generated response for the test set. Specifically, BLEU metric is to compare n-grams of the candidate with the n-grams of the reference text and count the number of matches. The more the matches, the better the candidate responses are. In this paper, we follow the prior work~\cite{zhang2018personalizing} and report the 1-gram BLEU scores.}

The statistics of all aforementioned datasets are listed in Table~\ref{tab:dataset}.

\begin{table*}[]
\centering
\caption{Results on GLUE benchmark. The results of comparison models are from correspondingly published papers. ``\textsc{Sf}" and ``\textsc{RA}" denote the shuffle and random noising functions, respectively, while ``\textsc{SfRa}" is the combination of both functions. In the ``SCORE", ``Avg'' means the average score in terms of the reported metrics, and ``$\Delta$'' refers to the performance improvements compared with baselines (marked with ``*''). {\zqh Best results are in bold.}}
\label{tab:main1}
\scalebox{1.08}{
\begin{tabular}{lcccccccccccccc}
\toprule
\multicolumn{1}{c}{\multirow{2}{*}{\textbf{Model}}} & \textbf{CoLA}          & \textbf{SST}      & \multicolumn{2}{c}{\textbf{MRPC}}              & \multicolumn{2}{c}{\textbf{STS-B}}             & \multicolumn{2}{c}{\textbf{QQP}}               & \multicolumn{2}{c}{\textbf{MNLI}}              & \textbf{QNLI}          & \textbf{RTE}          & \multicolumn{2}{c}{\textbf{SCORE}}      \\ \cmidrule{2-13}
\multicolumn{1}{c}{}                       & Mcc.           & Acc.           & F1    &Acc.             & Pcor.   &Scor.          & F1   &Acc.             & m.   &mm.               & Acc.           & Acc.           & Avg. &$\Delta$      \\ \midrule \midrule
\multicolumn{15}{l}{\textit{Single-task single models on dev set}}                                                              \\
BERT~\cite{devlin2019bert}          & 60.6  & 93.2  &   -                        & 88.0    & 90.0                   &  -                         & -                          & 91.3  & 86.6  &    -                       & 92.3  & 70.4  &- &-  \\
\textsc{UniLM}~\cite{dong2019unified}       & 61.1          & 94.5         & -    &-       &- &-    & -      & -         & 87.0 &85.9      & 92.7      & 70.9       &  -     &  -      \\
ERNIE2.0 \cite{sun2020ernie}     & 65.4  & 96.0    &  -                         & 89.7  & 92.3                 &  -                         &   -                        & \textbf{92.5}  & 89.1  &  -                         & 94.3  & 85.2 &- &-  \\
XLNet~\cite{yang2019xlnet}        & 63.6  & 95.6  &  -                         & 89.2  & 91.8                 &  -                         &   -                        & 91.8  & 89.8  &  -                         & 93.9  & 83.8 &- &- \\
T5-large~\cite{raffel2020exploring}      & 62.0 & 93.7 & \textbf{93.4} & 90.7 & 89.2          & 89.2 & 89.4 & 92.1 & 87.2 & 87.1 & 92.7 & 82.0 &- &- \\
% ELECTRA-large~\cite{clark2019electra} & 69.1  & 96.9  &  -         & 90.8  &- &92.6  &  -                         & 92.4  & 90.9  &  -                         & 95.0    & 88.0  &- &- \\
% T5 \cite{raffel2020exploring}                                        & 61.2          & 96.3          & 92.4   &89.9          & 89.9   &89.2          & 73.9   &89.9          & 89.9   &89.6          & 94.8          & 87.2          & 87.0  \\ 
\midrule
% & \multicolumn{9}{c}{\cellcolor{lightgray}\textit{Performance of baseline models}}                   \\ \cmidrule{2-10}
BART~\cite{lewis2020bart}   & 62.8          &\textbf{96.6}   &-   &90.4     &-   &91.2       &-   &\textbf{92.5}     & \textbf{89.9} &\textbf{90.1}      & \textbf{94.9}      & 87.0        & 88.4  &*   \\
\bartre                    & 63.6          & 96.4    &93.0      & 90.4     &90.6    & 90.8     &89.3    & 92.2      & \textbf{89.9} &89.7      & \textbf{94.9}      & 89.0        & 88.5    &+0.1   \\
\bartcont                  & 63.1          & 96.4     &93.2      & 90.6     &91.0     & 91.2     &89.7    & 92.3      & 89.8   &89.8      & 94.7      & 89.9      & 88.6   &+0.2   \\ \hdashline
%  & \multicolumn{9}{c}{\cellcolor{lightgray}\textit{Performance of E2S2-BART variants}}                                                                    \\ \cmidrule{2-10}
% \multicolumn{15}{c}{\textit{Performance of E2S2-BART variants on different combinations of denoising and contrastive objectives}}                 \\
\multicolumn{15}{l}{\textbf{E2S2-\bartde}} \\
\quad w/ \textsc{Sf}        & 63.9          & 96.0     &92.9         & 90.2     &\textbf{91.8}      & \textbf{92.0}    &89.7       & 92.3      & 89.6  &89.8      & 94.9      & 89.9      & 88.7    &+0.3 \\
\quad w/ \textsc{Ra}         & \textbf{66.0}         & 96.2         & 93.2       &\textbf{90.9}  &91.1 &91.4      & 89.8      & 92.4        & 89.6 &89.8     & 94.9   &89.5  &89.0 &+0.6  \\
\quad w/ \textsc{SfRa}       & 65.8          & 96.3     &93.2    & 90.7     &\textbf{91.8}   & 91.9     &\textbf{89.9}     & \textbf{92.5}      & 89.6 &89.8      & 94.8      & \textbf{91.0}  &\textbf{89.2}    &+0.8 \\
\textbf{E2S2-\bartjb} &64.6 &96.4 &93.2    &90.7 &91.7    &91.8 &89.8     &92.3  &89.8 &89.9 &94.8 &89.5 &88.9  &+0.5 \\
\textbf{E2S2-\bartjp} &65.3 &\textbf{96.6} &93.3  &\textbf{90.9} &91.7  &\textbf{92.0}   &89.8 &92.4   &89.7  &89.9   &\textbf{94.9} &90.3   &89.0  &+0.6 \\ \midrule \midrule
\multicolumn{15}{l}{\textit{Single-task single models on test set scored using the GLUE evaluation server}}                                                              \\
BERT~\cite{devlin2019bert}         & 60.5          & 94.9          & 89.3 &85.4          & 87.6   &86.5          & 72.1   &89.3          & 86.7   &85.9          & 92.7          & 70.1          & 83.4 &-      \\
\textsc{UniLM}~\cite{dong2019unified}    & 61.1              &  94.5             & 90.0 &-                   & -   &87.7                   &  71.7   &-                  & 87.0   &85.9                   & 92.7              &70.9               &83.7    &-        \\
Snorkel MeTa\cite{ratner2017snorkel}  &63.8 & 96.2  &91.5  &88.5   &90.1   &89.7  &73.1   &89.9      &87.6   &87.2      &93.9     &  80.8  & 86.0  &-  \\
MT-DNN++\cite{liu2019multi}  &\textbf{65.4} & 95.6  &91.1  &88.2   &89.6   &89.0  &72.7   &89.6      &87.9   &87.4      &\textbf{95.8}     &  85.1  & 86.5 &- \\
SemBERT~\cite{zhang2020semantics}      & 62.3                 & 94.6                 & 91.2                 & 88.3                 & 87.8                 & 86.7                 & 72.8                 & 89.8                 & 87.6                 & 86.3                 & 94.6                 & 84.5     &85.5 &-            \\
XLM~\cite{conneau2020unsupervised} & 62.9                 & 95.6                 & 90.7                 & 87.1                 & 88.8                 & 88.2                 & 73.2                 & 89.8                 & 89.1                 & 88.5                 & 94.0                   & 76.0     &85.3 &-              \\
SpanBERT~\cite{joshi2020spanbert}         &64.3 & 94.8  &90.9  &87.6   &89.9   &89.1  &71.8   &89.5      &88.1   &87.7      &94.3     &  79.0  & 85.6  &- \\
ERNIE2.0 \cite{sun2020ernie} & 63.5                 & 95.6                 & 90.2                 & 87.4                 & 91.2                 & \textbf{90.6}                 & 73.8                 & 90.1                 & 88.7                 & 88.8                 & 94.6                 & 80.2      &86.2 &-           \\
T5-large~\cite{raffel2020exploring} & 61.2                 & 96.3                 & \textbf{92.4}                 & \textbf{89.9}                 & \textbf{89.9}                 & 89.2                 & 73.9                 & 89.9                 & 89.9                 & \textbf{89.6}                 & 94.8                 & 87.2        &87.0 &-         \\
% T5 \cite{raffel2020exploring}                                        & 61.2          & 96.3          & 92.4   &89.9          & 89.9   &89.2          & 73.9   &89.9          & 89.9   &89.6          & 94.8          & 87.2          & 87.0  \\ 
\midrule
% & \multicolumn{9}{c}{\cellcolor{lightgray}\textit{Performance of baseline models}}                   \\ \cmidrule{2-10}
\bartre                                   & 61.7          & 96.4          & 91.1  &88.4          & 87.7   &87.0          & 73.0   &89.9          & 89.8   &89.0          & 94.6          & 86.5          & 86.2    &*   \\
\bartcont                               & 60.9          & 96.8          & 91.1  &88.5          & 88.6   &88.4          & 73.0   &89.9          & 89.9   &89.1          & 95.2          & 86.2          & 86.4   &+0.2   \\ \hdashline
%  & \multicolumn{9}{c}{\cellcolor{lightgray}\textit{Performance of E2S2-BART variants}}                                                                    \\ \cmidrule{2-10}
% \multicolumn{15}{c}{\textit{Performance of E2S2-BART variants on different combinations of denoising and contrastive objectives}}                 \\
\multicolumn{15}{l}{\textbf{E2S2-\bartde}} \\
\quad w/ \textsc{Sf}                                    & 60.1          & 96.5          & 91.5  &88.7          & 88.7   &88.5          & \textbf{74.4}   &\textbf{90.4} & 89.9   &89.0            & 95.1          & 86.7          & 86.6  &+0.4 \\
\quad w/ \textsc{Ra}                                      & 63.6          & 96.7          & 91.9  &89.1          & 88.8   &88.8          & 73.2   &89.8          & 90.0   &89.2          & 95.0          & 86.1          & 86.9  &+0.7 \\
\quad w/ \textsc{SfRa}                                 & 62.2          & \textbf{96.9} & 92.1  &89.4          & 88.7   &88.6          & 73.8   &90.3          & 90.0   &89.2          & 95.3 & 87.2          & 87.0  &+0.8 \\
\textbf{E2S2-\bartjb} &64.0 &\textbf{96.9}  &92.2  &89.5 &89.5   &89.2  &73.9   &90.2 &90.1   &89.3  &95.1 &87.3 &\textbf{87.3}  &+1.1 \\
\textbf{E2S2-\bartjp} &62.9 &96.8  &92.3  &89.7 &\textbf{89.9}   &89.8  &73.7   &90.1 &\textbf{90.2}   &89.1  &95.1 &\textbf{87.8} &\textbf{87.3}  &+1.1 \\

\bottomrule
\end{tabular}
}
\end{table*}

\subsection{Implementation Details}
As mentioned above, E2S2 is proposed for seq2seq transformer models. To validate the effectiveness of E2S2, we mainly implement it on a representative seq2seq model, \textit{i.e.}, BART~\cite{lewis2020bart}. For ease of illustration, we denote the BART trained with our E2S2 strategy as E2S2-BART.
Sequentially, we describe details of the pretraining and fine-tuning of E2S2-BART, respectively.

\subsubsection{\textbf{Pretraining}}
In practice, our model is based on the BART-large\footnote{\url{https://dl.fbaipublicfiles.com/fairseq/models/bart.large.tar.gz}} in the open-source toolkit fairseq\footnote{\url{https://github.com/pytorch/fairseq}}, with 24 transformer layers, a hidden size of 1024, a maximum sequence length of {\zqh 1024} and a learning rate of 1.5e-6. We initialize our E2S2-BART model with the corresponding pretrained BART model and continue pretraining the model for 50K update steps with a batch size of 2000. Following BART, we employ the same corpora for pretraining, including 160 GB of BookCorpus, Wiki, Stories, OpenWebText and CC-News. {\zqh During the data processing, we set the proportions of text infilling, Shuffle and Random as \{0.15, 0.05, 0.05\}, respectively. The span length of Shuffling is set to 3.}
All models are trained on the NVIDIA DGX SuperPOD cluster, in which each machine contains 8 x 40GB A100 GPUs. 

\subsubsection{\textbf{Fine-tuning}}
For the discriminative task, \ie GLUE benchmark, we follow the original BART and employ the final hidden state of the last decoder token as the final sentence representation and feed the hidden state into a multi-class linear classifier to obtain the final predictions. For RTE, STS-B and MRPC of GLUE benchmark, following previous works~\cite{liu2019roberta, he2020deberta}, we first finetune the pretrained E2S2-BART model on the MNLI dataset and then continue fine-tuning on their corresponding single-task corpus for better performance.  
As for \textsc{AbsSum}, GEC and \textsc{DiaGe} tasks, we input the source sentence (long document or grammatical-error sentence) into the encoder and generate the corresponding outputs on the decoder side auto-regressively. The detailed hyper-parameters of fine-tuning is shown in Table~\ref{tab:setting}. 
{\revision To avoid stochasticity, we report the average results over 5 random seeds for NLU tasks\footnote{{\revision Notably, due to the submission restrictions of GLUE test system, we select the models with the best validation results for each task and report their test results.}}, while for NLG tasks, we follow existing works~\cite{collins2005clause,ding2021progressive} and use the Bootstrap test~\cite{berg2012empirical} to calculate the statistical significance.}
% Appendix~\ref{appendix_parameters}.

\subsection{Compared Models}
We compare our E2S2-BART with the following models:
\begin{itemize}
\item \textbf{\bartre:} since the published paper of BART did not report the results on the GLUE leaderboard and GEC/\textsc{DiaGe} tasks, we reproduce the results of the original BART.
\item \textbf{\bartcont:} we initialize the model with the original BART and continually pretrain it with additional training steps, where the number of the additional training steps is equal to that of our E2S2-BART. This setting is to eliminate the doubt that the better performance of our E2S2 comes from more training steps.
\item \textbf{E2S2-\bartde:} one of E2S2-BART variant models trained with the proposed E2S2 strategy upon the single denoising objective. More specifically, we use the shuffle (denoted as \textsc{Sf}), random (denoted as \textsc{Ra}) replacement and their combination (denoted as \textsc{SfRa}) as extra noising functions, respectively.
\item \textbf{E2S2-\bartjb:} we simply combine the denoising and vanilla contrastive objectives, \ie the first solution for the contrastive objective that performs basic pooling operations to obtain sentence representations, and apply it to pretrain the E2S2-BART models.
\item \textbf{E2S2-\bartjp:} the combination of denoising and prompt-based contrastive objectives is adopted to guide the pretraining of E2S2-BART. Notably, prompt-based contrastive objective refers to the second solution that uses prompts to obtain sentence representations.

\end{itemize}

Moreover, for evaluating the effectiveness of our proposed method, we also compare against other competitive pretrained models, including BERT~\cite{devlin2019bert}, T5-large~\cite{raffel2020exploring}, UniLM~\cite{dong2019unified}, ERNIE2.0~\cite{sun2020ernie}, XLNet~\cite{yang2019xlnet}, SpanBERT~\cite{joshi2020spanbert}, SemBERT~\cite{zhang2020semantics}, etc. Note that the models with larger capabilities could always achieve higher performance, and thus we only compare our E2S2-BART models with the pretrained models of comparable size for a fair comparison. 

\begin{table*}
\centering
\caption{Experimental results on abstractive summarization task. Notably, for ease of illustration, we only apply the Shuffle-Random (w/ \textsc{SfRa}) noise as the denoising self-supervision on E2S2-based models. Best results are in bold. ``\sig'' indicates that E2S2-based models are significantly better than the baseline \bartcont at significance level $p<0.05$.}
\label{tab:main2}
\scalebox{1.0}{
\begin{tabular}{lccccccccc|ccc}
\toprule
\multicolumn{1}{c}{\multirow{2}{*}{\textbf{Model}}} & \multicolumn{3}{c}{\textbf{CNN/DM}} & \multicolumn{3}{c}{\textbf{XSum}}            & \multicolumn{3}{c}{\textbf{SAMSum}}     & \multicolumn{2}{c}{\textbf{Score}} \\ \cmidrule{2-12}
\multicolumn{1}{c}{}                       & Rouge\_1      & Rouge\_2      & Rouge\_L   & Rouge\_1      & Rouge\_2      & Rouge\_L    & Rouge\_1      & Rouge\_2      & Rouge\_L     & Avg.        & $\Delta$    \\ \midrule \midrule
%  & \multicolumn{9}{c}{\cellcolor{lightgray}\textit{Performance of baseline models}}                                                                                  \\ \cmidrule{2-10}
Lead-3       & 40.42 & 17.62 & 36.67 & 16.30  & 1.60   & 11.95 & 31.40  & 8.70   & 29.40  & -     & - \\
PTGEN~\cite{see2017get}        & 36.44 & 15.66 & 33.42 & 29.70  & 9.21  & 23.24 & 40.10  & 15.30  & 36.60 & -     & -  \\
BERTSumAbs~\cite{liu2019text}   & 41.72 & 19.39 & 38.76 & 38.86 & 16.33 & 31.15 & -     & -     & -    & -     & -  \\
RoBERTaShare~\cite{rothe2020leveraging} & 40.31 & 18.91 & 37.62 & 41.45 & 18.79 & 33.90  & -     & -     & -   & -     & -   \\
T5-large~\cite{raffel2020exploring}     & 41.74 & 19.66 & 39.14 & -     & -     & -     & 48.41 & 24.79 & \textbf{44.61} & -     & - \\
MASS~\cite{song2019mass}         & 42.12 & 19.50  & 39.01 & 39.75 & 17.24 & 31.95 & -     & -     & -    & -     & -  \\
UniLMv2~\cite{bao2020unilmv2}      & 43.16 & 20.42 & 40.14 & 44.00    & 21.11 & 36.08 & -     & -     & -  & -     & -   \\ 
\midrule
\bartre                                  & 43.88   & 21.09   & 40.64  & 44.49                & 21.81                & \multicolumn{1}{c}{36.69} &51.33 &26.63 &43.52  &36.68 &*  \\
\bartcont                                 & 43.90    & 21.00      & 40.63  & 45.19                & 21.80                 & \multicolumn{1}{c}{36.63} &51.72 &27.54 &43.94 &36.93 &+0.25   \\ \hdashline
% & \multicolumn{9}{c}{\cellcolor{lightgray}\textit{Performance of our E2S2-BART variant models}}                                                                      \\ \cmidrule{2-10} 
\textbf{E2S2-\bartde}                              & 44.12   & 21.23\sig   & 40.84  &\textbf{45.22}\sig               & 21.85                & \multicolumn{1}{c}{36.91\sig} &51.91\sig &27.47 &43.95  &37.07 &+0.39  \\
\textbf{E2S2-\bartjb}                         & 44.11   & 21.23\sig   & 40.87  & 44.87 & 21.84 & 36.88  &51.77 &27.39 &43.76    &36.97 &+0.29  \\
\textbf{E2S2-\bartjp}                 & \textbf{44.18}\sig   & \textbf{21.32}\sig   & \textbf{40.91}\sig  & 44.96 & \textbf{21.96}\sig & \textbf{36.95}\sig   &\textbf{52.06}\sig &\textbf{27.63} &44.12\sig   &\textbf{37.12} &+0.44  \\ \bottomrule
\end{tabular}
}
\end{table*}
\subsection{Main Results}
We first present the detailed results of our E2S2-BART variant models and other cutting-edge pretrained models on the GLUE benchmarks, and then compare our models with the baselines on the generative summarization, GEC and dialogue generation tasks for further analysis.

\subsubsection{GLUE Results} We follow the previous work~\cite{liu2019roberta} and report results on both dev and test sets of GLUE benchmark in Table~\ref{tab:main1}, where the results of test sets are obtained from the GLUE leaderboard.
Compared with the powerful pretrained models and baseline models, our E2S2-BART models consistently achieve better performance on most tasks of GLUE benchmark, indicating the effectiveness of our E2S2 method. In particular, when only using the denoising objective, it can be seen that Shuffle+Random (``w/ \textsc{SfRa}'') objective generally performs better than the single Shuffle or Random objective, and achieves an average score of 89.2\% and 87.0\% on dev and test sets, respectively. More specifically, for the CoLA task, the Random objective significantly outperforms the shuffle objective, indicating that denoising the random replacement could be more beneficial to the correction of grammatical errors in CoLA dataset. 

In addition, E2S2-\bartjb and E2S2-\bartjp outperform all compared models. Compared with the original BART, the averaged performance improvement on the test sets in improved by  +1.1\%. These results prove that more self-supervised information on the encoder side can improve the ability of language understanding, confirming our claim. To be more specific, from the test results of our E2S2 variants, we find that the models with contrastive learning can achieve higher performance on STS-B task, which is widely used to evaluate the performance of sentence representations \cite{gao2021simcse,yan2021consert}. This result shows that contrastive learning could alleviate the problem of anisotropy and make the encoder output more efficient sentence representations, thus benefiting the understanding of sentences. Moreover, prompt-based contrastive learning can achieve further improvements, indicating that a prompt-based approach can indeed obtain more reliable and informative sentence representations. 

\begin{figure}[t]
\centering
\includegraphics[width=0.4\textwidth]{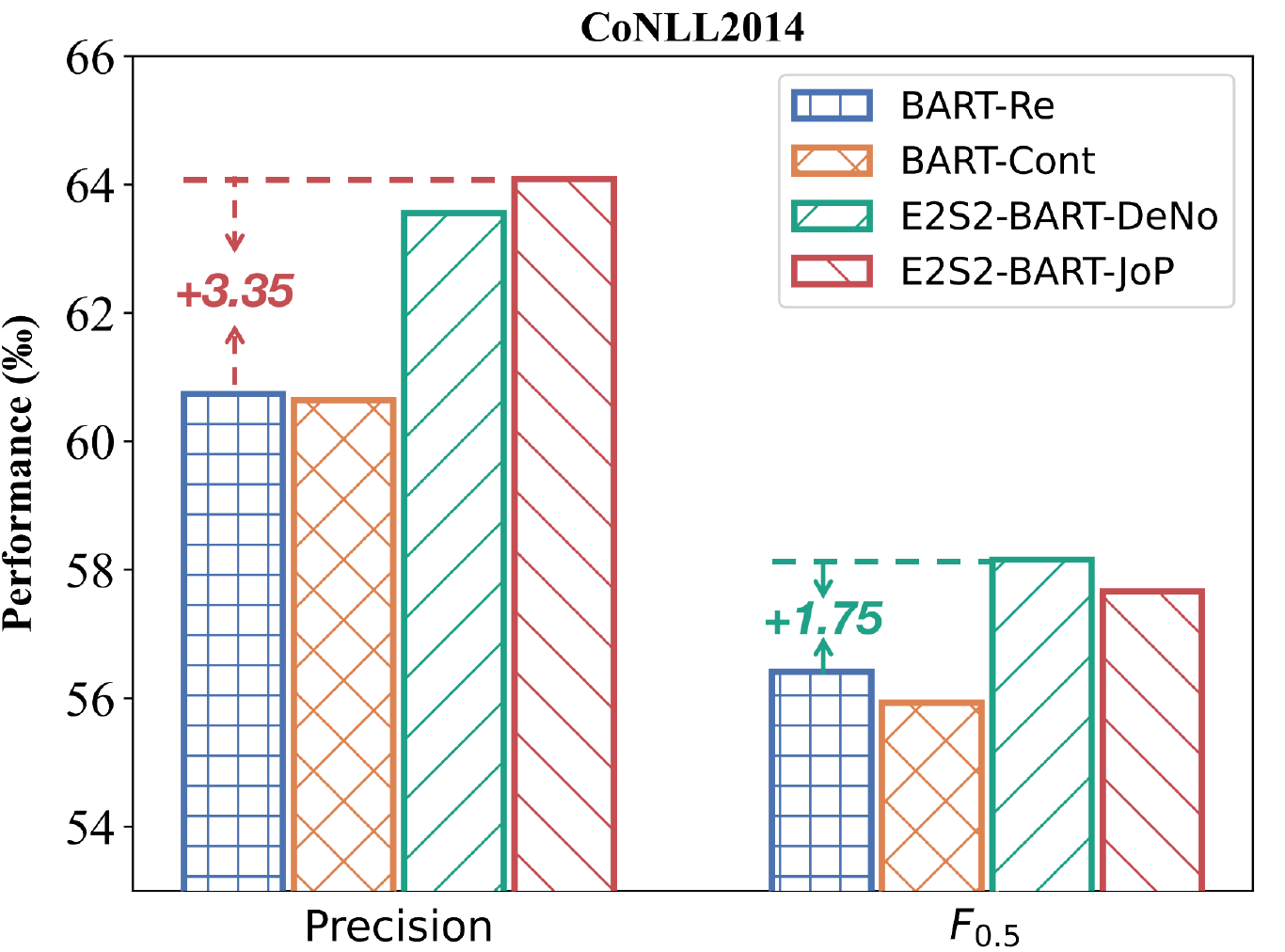} 
    % \centering
	\caption{Results on GEC task (CoNLL2014).}
	\label{e2s2:fig3}
\end{figure}

\subsubsection{Abstractive Summarization Results}
Table~\ref{tab:main2} reports the comparison results of CNN/DM, XSum and SAMSum datasets. Similar to the setting of Table~\ref{tab:main1}, we also divide the results into several groups. The first group lists the results of some powerful generative methods, and the second group reports the results of our E2S2-BART variants. Regarding the denoising objectives of E2S2-BART variants, we only present the results of ``w/ \textsc{SfRa}'' setting for ease of illustration. 

As shown in Table~\ref{tab:main2}, with the help of our proposed pretraining strategy, E2S2-BART models are superior to other cutting-edge models, and substantially outperform the original BART model and its continual pretrained version on both abstractive summarization tasks. These results further prove that our E2S2 models can also perform better on the tasks of language generation. Notably, the improvement of E2S2 on abstractive summarization tasks is slightly smaller than that on the GLUE benchmark. One possible reason is that the text summarization task is not sensitive to the noise of input sentences, and more independent on the syntactic and semantic knowledge of pretrained models. As claimed in Krishna~et~al.~\cite{krishna2021does}, for text summarization task, the models pretrained with nonsense words can still achieve comparable performance with the carefully pretrained models.

\subsubsection{GEC Results} We also compare E2S2-\bartde and E2S2-\bartjp with the baselines, \textit{i.e.}, \bartre and \bartcont, and show the results in Fig.~\ref{e2s2:fig3}. Similarly, under the denoising self-supervision, our E2S2 method (\textit{Precision: 63.56\%, $F_{0.5}$: 58.16\%}) achieves a significant performance improvement compared with \bartre (\textit{Precision: 60.74\%, $F_{0.5}$: 56.41\%}). This is because that the denoising pretraining objectives are strongly relative to the task of grammatical error correction. Interestingly, different from the other two tasks, \textit{i.e.}, GLUE benchmark and abstractive summarization, the prompt-based E2S2-\bartjp (\textit{Precision: 64.09\%, $F_{0.5}$: 57.66\%}) performs slightly worse on the GEC task. One possible reason is that the prompt template itself could introduce some noise and thus affect the ability of model on noise detection. Additionally, compared with the original BART, the continual pretrained BART achieves suboptimal performance on all metrics, which indicates that a longer pretraining could not always be better and continues proving the effectiveness of E2S2.

\begin{table*}
    \centering
    \caption{Experimental results on dialogue generation task.  The lower (``$\downarrow$'') PPL and higher (``$\uparrow$'') BLEU scores refer to better performance. ``\sig'' indicates that E2S2-based models are significantly better than the baseline \bartcont at significance level $p<0.05$.}
    \label{tab:main3}
    \begin{tabular}{lcccccccccc}
    \toprule    
    \multicolumn{1}{c}{\multirow{2}{*}{\textbf{Model}}} & \multicolumn{2}{c}{\textbf{\pchat}} & \multicolumn{2}{c}{\textbf{\daily}}            & \multicolumn{2}{c}{\textbf{DSTC7}}  & \multicolumn{2}{c}{\textbf{\echat}}   & \multicolumn{2}{c}{$\Delta$} \\ \cmidrule(lr){2-3} \cmidrule(lr){4-5} \cmidrule(lr){6-7} \cmidrule(lr){8-9} \cmidrule(lr){10-11} 
    \multicolumn{1}{c}{}                       &PPL ($\downarrow$) &\revision BLEU ($\uparrow$)     &PPL ($\downarrow$) &\revision BLEU ($\uparrow$)     &PPL ($\downarrow$)  &\revision BLEU ($\uparrow$)    &PPL ($\downarrow$)  &\revision BLEU ($\uparrow$)  
    &PPL ($\downarrow$)  &\revision BLEU ($\uparrow$) \\ \midrule 
    %  & \multicolumn{9}{c}{\cellcolor{lightgray}\textit{Performance of baseline models}}                                                                                  \\ \cmidrule{2-10}		
    \bartre                                  & 18.90 &\revision 28.12  & 17.88   &\revision 27.10  & 8.81  &\revision 39.37  & 26.36    &\revision 19.84                     & *   &\revision *  \\
    \bartcont         & 15.89  &\revision 28.20   & 17.18  &\revision 27.43   & 8.37  &\revision 39.84  & 26.01         & \revision 19.22          & 1.13  &\revision 0.06  \\ \hdashline
    E2S2-\bartde         & 15.72  &\revision 28.26  & \textbf{16.60}  &\revision \bf  27.58\sig   & 8.17  & \revision \bf 39.98\sig & 25.58    &  \revision 20.87\sig      & 1.47   & \revision 0.56  \\
    E2S2-\bartjb                                        & \textbf{15.48}  & \revision \bf 28.33\sig   & 16.67  &\revision 27.53   & 8.20  &\revision 39.97  & 25.27   &  \revision 21.07\sig                  & 1.58   & \revision \bf 0.62 \\
    E2S2-\bartjp                                & 15.52  &\revision 28.32  & \textbf{16.60}   &\revision 27.48  & \textbf{8.09}  &\revision 39.90  & \textbf{25.23}             &  \revision \bf 21.12\sig         & \textbf{1.63}  & \revision 0.59  \\ \bottomrule
    \end{tabular}
\end{table*}
\subsubsection{Dialogue Generation Results} Table~\ref{tab:main3} lists the results of our models and other baselines on several dialogue generation datasets. Obviously, our E2S2 models improve significant performance on these dialogue datasets. Specifically, compared with \bartre, our E2S2-\bartjb decreases the validation PPL by 3.42 on \pchat dataset, {\revision and our E2S2-\bartjp brings 1.28 BLEU gains on \echat dataset}. 

The aforementioned results on both discriminative and generative tasks prove that the encoding-enhanced strategy not only improves the language understanding performance (more dependent on the abilities of the encoder), but also be helpful to the decoder-side language generation tasks via providing better sentence representation from the encoder.

\subsection{Ablation Study}
We conduct extensive ablation studies to investigate the effectiveness of multiple pretraining objectives in E2S2, {\zqh the effect of coefficient $\lambda_{de}$ and $\lambda_{cl}$,} and analyze the influence of different designed prompt templates used in prompt-based contrastive learning. Notably, some representative tasks of GLUE benchmark, \textit{e.g.}, CoLA, MRPC and STS-B, with their original training and dev sets are used for our analysis. We apply the \textit{Shuffle}-\textit{Random} noising function as the denoising self-supervision for all experiments.

\begin{table}[t]
\centering
\caption{Analysis of different pretraining objectives used in E2S2 ($\{\mathcal{L}^*_{nll},\mathcal{L}_{nll},\mathcal{L}_{de},\mathcal{L}_{cl}\}$) , evaluated on some representative tasks.}
\label{tab:ablation1}
\begin{tabular}{c|ccccccc}
\toprule
& $\mathcal{L}^*_{nll}$ & $\mathcal{L}_{nll}$ &$\mathcal{L}_{de}$ &$\mathcal{L}_{cl}$ & \textbf{CoLA} & \textbf{MRPC} & \textbf{STS-B}\\ \midrule
M1  &\checkmark &    &    &    & 60.4     & 90.3/87.7     &  87.4/87.0        \\
M2  &\checkmark &    &\checkmark  &    &  63.6    & 91.9/89.1     &88.8/88.8         \\
M3 &\checkmark &    &\checkmark  &\checkmark  &  61.6    & 90.6/88.1     &89.0/88.7          \\
M4 &\checkmark &\checkmark  &    &\checkmark  & 61.1     &  91.1/88.5    &89.5/88.9          \\
M5 &\checkmark &\checkmark  &\checkmark  &\checkmark  &\textbf{64.0}      & \textbf{92.2/89.5}     & \textbf{89.5/89.2}         \\ \bottomrule
\end{tabular}
\end{table}

\subsubsection{\textbf{Influence of the Pretraining Objectives}}
As Eq.~\ref{eq4}, we optimize E2S2 with the combination of multiple training objectives, \textit{i.e.}, $\{\mathcal{L}^*_{nll},\mathcal{L}_{nll},\mathcal{L}_{de},\mathcal{L}_{cl}\}$. To investigate their effectiveness, we report the results of different objective combinations in Table~\ref{tab:ablation1}. Notably, to avoid the influence of prompt, we did not use the prompt-based approach in this study. For clarity, the models are denoted as M1 to M5 from top to bottom.

As shown in Table~\ref{tab:ablation1}, the model with all pretraining objectives (\textit{i.e.}, M5) performs best and all objectives are beneficial to our E2S2 model consistently. Specifically, with the help of $\mathcal{L}_{de}$, M2 greatly improves over the vanilla seq2seq model M1, especially on the CoLA task, with a improvement of +3.2\%. Similarly, when using the full contrastive objective, \textit{i.e.}, $\{\mathcal{L}^*_{nll},\mathcal{L}_{nll},\mathcal{L}_{cl}\}$, M4 outperforms M1 consistently and achieves +2.0\% averaged gains in terms of Pearson-Spearman correlations on STS-B task. These results prove the significance of integrating more self-supervised objectives on the encoder side of seq2seq pretrained models, which confirms our claim. 

\subsubsection{{\zqh \textbf{Parameter analyses of $\lambda_{de}$ and $\lambda_{cl}$}}}
\label{sec:parameter_analysis}
{\zqh The factors $\lambda_{de}$ and $\lambda_{cl}$ in Eq.~\ref{eq4}, which are used to balance different objectives, are two important hyper-parameters in our E2S2. Here, we analyze their influence by evaluating the performance of BART-large with different $\lambda_{de}$ and $\lambda_{cl}$ spanning \{0.05, 0.1, 0.5\}. Fig.~\ref{e2s2:fig4} illustrates the average dev results of the GLUE benchmark. 

\begin{figure}[t]
\includegraphics[width=0.45\textwidth]{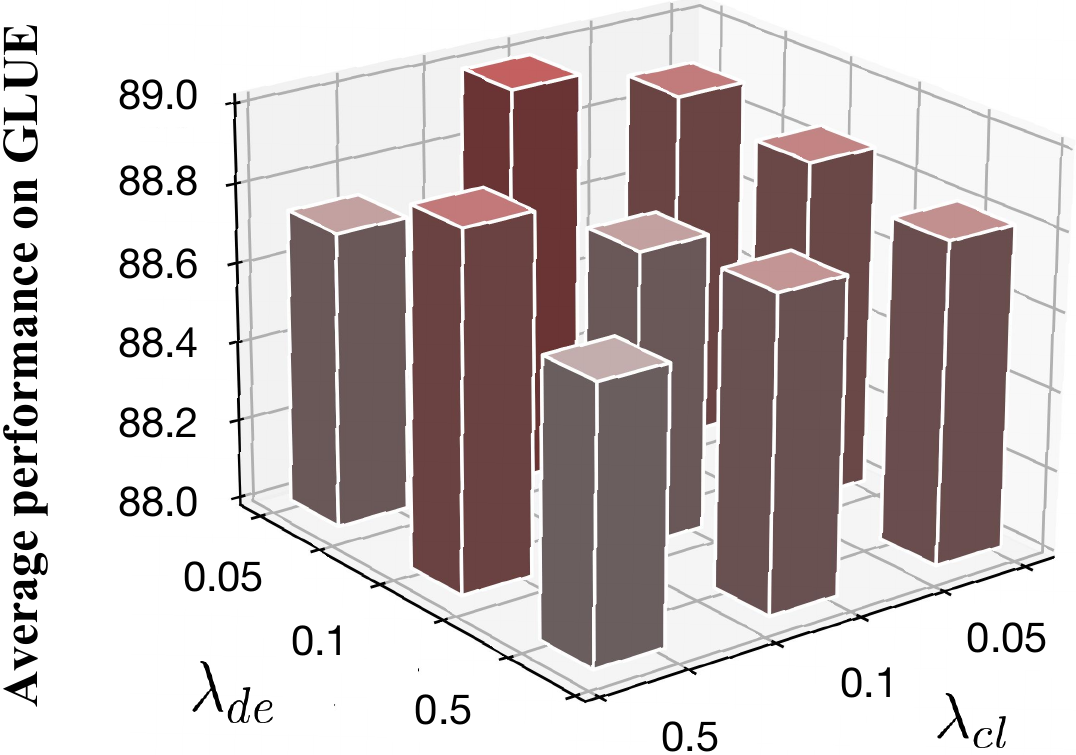} 
    % \centering
	\caption{{\zqh Parameter analyses of $\lambda_{de}$ and $\lambda_{cl}$. We train the E2S2-\bartjb models with different coefficient combinations, and evaluate on the dev sets of GLUE benchmark.}}
	\label{e2s2:fig4}
\end{figure}

As seen, compared with the baseline, \textit{i.e.}, \bartcont with a 88.6 average score, our E2S2-BART models consistently achieve better performance across all ratios of $\lambda_{de}$ and $\lambda_{cl}$, basically indicating that the performance of E2S2 is not sensitive to these factors. More specifically, the case of $\lambda_{de} = 0.05$ and $\lambda_{cl} = 0.1$ performs best, and we thereby use this setting in our experiments.
}

\subsubsection{\textbf{Effectiveness of Prompt Learning}}
\label{prompt}
As mentioned above, the design of prompt template is the key challenge of prompt-based learning~\cite{liu2021pre}. To further analyze the influence of different prompts for our E2S2 models, we conduct comparative experiments on 6 manual templates. The prompt templates and their corresponding results are shown in Table~\ref{tab:ablation2}. 

\begin{table}[t]
\centering
\caption{Analysis of different prompt templates, which are indexed from T1 to T6. In addition to the representative tasks of GLUE, we also evaluate the models on GEC task.}
\label{tab:ablation2}
\scalebox{0.95}{
\begin{tabular}{cccccccc}
% \toprule
\toprule
\textbf{Index} & \multicolumn{7}{l}{\textbf{Template}}                                 \\ \midrule
T1    & \multicolumn{6}{l}{\texttt{[MASK]} means \texttt{[X]}.}                                     \\
T2    & \multicolumn{6}{l}{\texttt{[MASK]} represents \texttt{[X]}.}                                \\
T3    & \multicolumn{6}{l}{\texttt{[MASK]} means the sentence : ``\texttt{[X]}''.}                    \\
T6    & \multicolumn{6}{l}{\texttt{[MASK]} represents the sentence : ``\texttt{[X]}''.}               \\
T5    & \multicolumn{6}{l}{\texttt{[MASK]} is the meaning of the sentence : ``\texttt{[X]}''.}        \\
T6    & \multicolumn{6}{l}{\texttt{[MASK]} is the representation of the sentence : ``\texttt{[X]}''.} \\ \midrule \midrule
\multirow{2}{*}{\textbf{Index}} & \textbf{CoLA} & \multicolumn{2}{c}{\textbf{MRPC}}   & \multicolumn{2}{c}{\textbf{STS-B}}     & \multicolumn{2}{c}{\textbf{CoNLL2014}} \\ \cmidrule{2-8} 
                       & Mcc.  & F1 &Acc. & Pcor. &Scor. &Precision & $F_{0.5}$ \\ \midrule
T1       & 57.4     &  91.2 &88.2      & 88.4 &78.9     & 59.57 & 55.88    \\
T2       & 59.7    &  91.8 &\textbf{89.3}      &  89.4 &89.2     & 59.18    & 55.94    \\
T3       & 61.9     & 91.4 &88.8       & 89.3 &89.2     &59.23 &55.32     \\
T4       &  60.9    &  92.0 &89.2      & 88.9 &88.8     &58.92   & 55.28    \\
T5       & 62.8     & 91.8 &89.0       & 89.2 &89.0     & 61.23  &56.88     \\
T6       & \textbf{62.9}     & \textbf{91.9} &89.1       &  \textbf{89.8} &\textbf{89.5}     & \textbf{64.09}   &\textbf{57.66}     \\ \bottomrule
\end{tabular}
}
\end{table}

Similar to the previous works~\cite{shin2020autoprompt,jiang2022promptbert}, different prompt templates achieve varied performance. In particular, the simple templates, \textit{e.g} ``\texttt{[MASK]} means \texttt{[X]}.", perform worse than the other sophisticated templates, indicating the importance of manual template engineering. More specifically, the last prompt T6 performs best among all designed prompts, thus adopting it as the default setting. It is also noteworthy that manual template engineering is usually time-consuming and may fail to design optimal prompts, as the above prompts mostly fall short in the GEC task. The automated template learning, which can create the actual text prompts or continuous soft prompts automatically, could address the above problems efficiently \cite{lester2021power,shin2020autoprompt}. More potential prompt engineering will be explored in our future works.

\begin{table}[t]
    \centering
    \caption{Results of the original T5-base model~\cite{raffel2020exploring} and those equipped with our E2S2, which show that T5-base can also benefit from E2S2 strategy, proving the compatibility of E2S2 with other cutting-edge PLMs.}
    \label{tab:analysis1}
    \begin{tabular}{lcccccc}
    \toprule
    \multicolumn{1}{c}{\multirow{2}{*}{Method}} & SST-2         & MRPC          & STS-B          & QNLI          & \multicolumn{2}{c}{MNLI}        \\ \cmidrule{2-7}
    \multicolumn{1}{c}{}                        & Acc. & Acc. & Pcor. & Acc. & m. & mm.    \\ \midrule
    T5-base~\cite{raffel2020exploring}                                    & 92.7          & 88.9          & 88.0           & 90.5          & 84.2      &84.6    \\
    \tre                                       & 93.7          & 88.7          & 88.9           & 92.5          & 87.0   &86.7       \\
    \tcont                                     & 93.9          & 88.2          & 89.7           & 92.9          & 87.2   &86.6       \\
    E2S2-\tde                                      & 94.0            & 89.2          & 89.7           & 92.9          & 87.1    &\textbf{87.3}      \\
    E2S2-\tjb                                      & \textbf{94.5} & 90.2          & \textbf{90.3}  & 92.9          & \textbf{87.3}     &87.0     \\
    E2S2-\tjp                                       & 94.3          & \textbf{90.7} & 90.0           & \textbf{93.1} & \textbf{87.3} &87.2 \\
    \bottomrule
    \end{tabular}
\end{table}

\begin{table*}[]
\centering
\caption{\zqh Results of BART-base models on the dev sets of GLUE benchmark. Notably, all models are trained from scratch for 5 epochs. In the E2S2-\bartde setting, we use the ``\textsc{SfRa}'' as extra noising functions.}
\label{tab:training_from_scratch}
\scalebox{1.08}{
\begin{tabular}{lcccccccccccccc}
\toprule
\multicolumn{1}{c}{\multirow{2}{*}{\textbf{Model}}} & \textbf{CoLA}          & \textbf{SST}      & \multicolumn{2}{c}{\textbf{MRPC}}              & \multicolumn{2}{c}{\textbf{STS-B}}             & \multicolumn{2}{c}{\textbf{QQP}}               & \multicolumn{2}{c}{\textbf{MNLI}}              & \textbf{QNLI}          & \textbf{RTE}          & \multicolumn{2}{c}{\textbf{SCORE}}      \\ \cmidrule{2-13}
\multicolumn{1}{c}{}                       & Mcc.           & Acc.           & F1    &Acc.             & Pcor.   &Scor.          & F1   &Acc.             & m.   &mm.               & Acc.           & Acc.           & Avg. &$\Delta$      \\ \midrule \midrule
\multicolumn{15}{l}{\textit{Single-task single models on dev set}}                                                              \\
BART-base (Vanilla)          & 32.3 & 88.5 & 88.2 & 83.3 & 81.5 & 81.9 & 85.2 & 89.1 & 75.8 & 77.0 & 86.5 & 69.3 & 76.0 & *  \\
\textbf{E2S2-\bartde}          & 38.2 & 88.5 & \textbf{89.3} & \textbf{85.5} & 84.6 & \textbf{84.4} & \textbf{86.0} & \textbf{89.8} & \textbf{76.8} & 76.9 & 87.3 & 70.0 & 77.6 & +1.6  \\
\textbf{E2S2-\bartjb}          & \textbf{38.5} & \textbf{88.8} & \textbf{89.3} & 84.8 & \textbf{84.8} & \textbf{84.4} & 85.5 & 89.5 & 76.7 & \textbf{77.9} & \textbf{88.4} & \textbf{70.8} & \textbf{77.9} & +1.9  \\
\bottomrule
\end{tabular}
}
\end{table*}

\subsection{Discussion and Analysis}
\label{disscusion}
\subsubsection{\textbf{Compatibility with other seq2seq PLMs}} As aforementioned, we show the effectiveness of our E2S2 strategy on BART model. To further prove the universality of our proposed strategy, we examine whether the strategy is compatible with other seq2seq PLMs. Specifically, we apply E2S2 strategy to another popular seq2seq PLM, \textit{i.e.}, T5-base~\cite{raffel2020exploring}. Similar to the pretraining settings in E2S2-BART, we continue pretraining the original T5 model using E2S2-variant strategies\footnote{{\zqh Apart from our E2S2 method, we basically follow the pretraining processes in the original paper~\cite{raffel2020exploring}. For example, we replace the spans of corrupted tokens with sentinel tokens, and enforce the decoder to generate the output sequence that consists of the dropped-out spans, delimited by the sentinel tokens used to replace them in the input. You can refer to the original paper~\cite{raffel2020exploring} for more details.}} and denote the obtained models as ``E2S2-T5\_*''. E2S2-based T5 models are evaluated on several GLUE sub-tasks and the results are listed in Table~\ref{tab:analysis1}.

Compared with the baselines T5\_\textsc{Re} and T5\_\textsc{Cont}, our E2S2-T5 models achieve consistent performance improvements on these tasks, where the improvements on MRPC and STS-B tasks are +2.5\% and +1.4\%, respectively. These results demonstrate that our E2S2 is not only beneficial to BART model, but also works well on T5 model. \textbf{Takeaway}: \textit{Our proposed E2S2 strategy is universal and can be applied to more seq2seq PLMs.}

\subsubsection{\zqh \textbf{Is E2S2 still helpful when training from scratch?}}
{\zqh
In the above experiments, we trained the E2S2-BART/T5 models in the continual pretraining manner, \textit{i.e.}, based on a warm start from BART/T5. Some readers may concern that adding objectives in the E2S2 could affect the stability of regular pretraining and may wonder whether our E2S2 still be helpful when training from scratch. To address this concern, we further adopt our E2S2 into the regular pretraining phase (started from random weights) of BART-base\footnote{\zqh Due to the limited computational budget, we conduct experiments on the BART-base models trained with different methods for 5 epochs. The pretraining hyper-parameters are similar to those in the original paper~\cite{lewis2020bart}.}, and report the contrastive results in Table~\ref{tab:training_from_scratch}. Notably, we do not adopt the E2S2-\bartjb here, as our focus is to investigate the influence of adding objectives (denoising and contrastive objectives), rather than the prompting method.

As seen, compared to the vanilla pretraining method, both of our E2S2 methods bring consistent and significant performance improvements, proving that adding our proposed objectives in the regular pretraining phase is still helpful. We attribute this to the complementarity between these adding objectives and original reconstruction objective in BART~\cite{lewis2020bart}, as our E2S2 aims to encourage the encoder to learn better representations that are helpful for the decoder to reconstruct the sentences. \textbf{Takeaway}: \textit{Our E2S2 is still helpful when training the models from scratch.}
}

\subsubsection{\textbf{Are the Encoder Indeed Enhanced?}}
As stated in section~\ref{intro}, the goal of E2S2 is to enhance the encoder of seq2seq models, so that the encoder can provide more discriminative representations and be beneficial to the understanding and generation of natural language. To explore whether the goal is achieved, 
%we present a detailed analysis and discussion. Specifically, 
we conduct experiments on the encoder itself with the probing tasks \cite{conneau2018senteval} to evaluate the effect of learning representations of the encoder. The probing tasks aim to study the simple linguistic properties of sentences, which can be classified into three groups: a)  ``Surface": used to evaluate the simple surface properties; b) ``Syntactic": used to quantify the syntactic reservation ability; c) ``Semantic": used to analyze the deeper semantic representation ability. Following Hao~et~al.~\cite{hao2019modeling}, we introduce an MLP classifier on the encoder side and train the classifier on the train sets of probing tasks. {\zqh For reference, we also compare our E2S2 method with the other two widely-used representation learning methods, \textit{i.e.}, SimCSE~\cite{gao2021simcse} and PromptBERT~\cite{jiang2022promptbert}. In practice, we basically follow the unsupervised training processes in the original papers to improve the encoder of BART, except that we use the average output of the last hidden layer as the sentence representation during the SimCSE training.
}

The contrastive results on dev sets of these probing tasks are listed in Table~\ref{tab:analysis2}. As seen, compared with the vanilla seq2seq pretraining, our E2S2 method achieves better performance on all probing tasks, with an averaged gain of +1.25\%. {\zqh Moreover, we can see that our E2S2 achieves comparable or even better performance against the powerful SimCSE and PromptBERT methods that have been widely proved to effectively improve the sentence representations of encoder. These results can prove the superiority of our method.}
\textbf{Takeaway}: \textit{The proposed E2S2 strategy enables the encoder to learn better sentence representations that preserve more surface, syntactic and semantic knowledge.}

\begin{table}[]
\centering
\caption{Results on 10 probing tasks that are used to evaluate sentence representations of the encoder output.}
\label{tab:analysis2}
\begin{tabular}{ccccccc}
\toprule
\multicolumn{2}{c}{\textbf{Task}}          & \textbf{vanilla}  & \textbf{\zqh SimCSE}  & \textbf{\zqh PromptBERT} & \textbf{E2S2}                     \\ \midrule \midrule	
\multirow{2}{*}{\textbf{Surface}}   & SeLen & 66.1    &\textbf{\zqh 68.6} &\zqh 64.9   & 68.0                            \\
                           & WC    & 57.5  &\zqh 57.6 &\zqh 57.5    &  \textbf{59.4}                            \\ \midrule
\multirow{3}{*}{\textbf{Syntactic}} & TrDep & 17.4  &\textbf{\zqh 20.6}  &\zqh 18.1    & 17.8                          \\
                           & ToCo  &  10.7  &\textbf{\zqh 13.2} &\zqh 12.3    &  11.4                         \\
                           & BShif & 79.6  &\zqh 76.4 &\zqh 77.7  &  \textbf{80.2}                           \\ \midrule
\multirow{5}{*}{\textbf{Semantic}}  & Tense &  64.6  &\zqh 67.5  &\textbf{\zqh 68.4}   & 65.5                         \\
                           & SubN  &  50.5  &\textbf{\zqh 53.8}  &\zqh 51.8   & 51.0                           \\
                           & ObjN  & 55.8 &\textbf{\zqh 57.8}  &\zqh 56.5    &  56.1                          \\
                           & SoMo  & 49.9 &\zqh 47.5  &\zqh 47.5    &  \textbf{55.3}                          \\
                           & CoIn  &  66.9  &\zqh 66.9   &\zqh 66.9  & \textbf{67.1}                            \\ \bottomrule
\end{tabular}
\end{table}

\section{Conclusion}
\label{sec:conclusion}
In this paper, we propose an encoding-enhanced seq2seq pretraining strategy (E2S2) to improve the vanilla seq2seq pretrained models. Instead of only optimizing the text infilling objective on the decoder side, E2S2 presents two self-supervised pretraining objectives in a local-to-global manner, \textit{i.e.}, locally denoising the corrupted sentences and globally learning better sentence representations, on the encoder side to provide richer supervisions. With the help of these self-supervised information, the encoder is able to effectively distinguish the noise tokens and capture high-level (\textit{i.e.}, syntactic and semantic) knowledge, thus boosting the performance of the seq2seq model on both language understanding and generation tasks. Experiments show that E2S2 improves the seq2seq models on several tasks consistently, especially on the GLUE benchmarks and grammatical error correction. These results demonstrate the effectiveness and universality of our E2S2. 

Future work includes validating the E2S2 on larger seq2seq PLMs\footnote{\zqh Considering that our E2S2 is a plug-and-play method, we believe that applying our method to larger cutting-edge models has the potential to achieve much better performance on the GLUE benchmark. However, due to the limited compute resources, we do not experiment on these large language models in this paper.}, \textit{e.g.}, T5-11B~\cite{raffel2020exploring}, and integrating the automated prompt template~\cite{shin2020autoprompt} or soft prompt learning~\cite{lester2021power} to our E2S2 method. Additionally, it is also interesting to validate the effectiveness of E2S2 on more challenging downstream tasks, \textit{e.g.}, translation, in the future. Our work provides a new view of the seq2seq language model pretraining and we hope it can foster future self-supervision research in this field.

\bibliographystyle{IEEEtran}
\bibliography{tkde.bib}

\begin{thebibliography}{10}
\providecommand{\url}[1]{#1}
\csname url@samestyle\endcsname
\providecommand{\newblock}{\relax}
\providecommand{\bibinfo}[2]{#2}
\providecommand{\BIBentrySTDinterwordspacing}{\spaceskip=0pt\relax}
\providecommand{\BIBentryALTinterwordstretchfactor}{4}
\providecommand{\BIBentryALTinterwordspacing}{\spaceskip=\fontdimen2\font plus
\BIBentryALTinterwordstretchfactor\fontdimen3\font minus
  \fontdimen4\font\relax}
\providecommand{\BIBforeignlanguage}[2]{{%
\expandafter\ifx\csname l@#1\endcsname\relax
\typeout{** WARNING: IEEEtran.bst: No hyphenation pattern has been}%
\typeout{** loaded for the language `#1'. Using the pattern for}%
\typeout{** the default language instead.}%
\else
\language=\csname l@#1\endcsname
\fi
#2}}
\providecommand{\BIBdecl}{\relax}
\BIBdecl

\bibitem{ramachandran2017unsupervised}
P.~Ramachandran, P.~J. Liu, and Q.~Le, ``Unsupervised pretraining for sequence
  to sequence learning,'' in \emph{EMNLP}, 2017.

\bibitem{song2019mass}
K.~Song, X.~Tan, T.~Qin, J.~Lu, and T.-Y. Liu, ``Mass: Masked sequence to
  sequence pre-training for language generation,'' in \emph{ICML}, 2019.

\bibitem{lewis2020bart}
M.~Lewis, Y.~Liu, N.~Goyal, M.~Ghazvininejad, A.~Mohamed, O.~Levy, V.~Stoyanov,
  and L.~Zettlemoyer, ``Bart: Denoising sequence-to-sequence pre-training for
  natural language generation, translation, and comprehension,'' in \emph{ACL},
  2020.

\bibitem{raffel2020exploring}
C.~Raffel, N.~Shazeer, A.~Roberts, K.~Lee, S.~Narang, M.~Matena, Y.~Zhou,
  W.~Li, and P.~J. Liu, ``Exploring the limits of transfer learning with a
  unified text-to-text transformer,'' \emph{JMLR}, 2020.

\bibitem{qi2020prophetnet}
W.~Qi, Y.~Yan, Y.~Gong, D.~Liu, N.~Duan, J.~Chen, R.~Zhang, and M.~Zhou,
  ``Prophetnet: Predicting future n-gram for
  sequence-to-sequencepre-training,'' in \emph{Findings of EMNLP}, 2020.

\bibitem{liu2020multilingual}
Y.~Liu, J.~Gu, N.~Goyal, X.~Li, S.~Edunov, M.~Ghazvininejad, M.~Lewis, and
  L.~Zettlemoyer, ``Multilingual denoising pre-training for neural machine
  translation,'' \emph{TACL}, 2020.

\bibitem{lewis2020pre}
M.~Lewis, M.~Ghazvininejad, G.~Ghosh, A.~Aghajanyan, S.~Wang, and
  L.~Zettlemoyer, ``Pre-training via paraphrasing,'' in \emph{NeurIPS}, 2020.

\bibitem{wang2019denoising}
L.~Wang, W.~Zhao, R.~Jia, S.~Li, and J.~Liu, ``Denoising based
  sequence-to-sequence pre-training for text generation,'' in \emph{EMNLP},
  2019.

\bibitem{zhou2021improving}
W.~Zhou, T.~Ge, C.~Xu, K.~Xu, and F.~Wei, ``Improving sequence-to-sequence
  pre-training via sequence span rewriting,'' in \emph{EMNLP}, 2021.

\bibitem{li2020survey}
J.~Li, A.~Sun, J.~Han, and C.~Li, ``A survey on deep learning for named entity
  recognition,'' \emph{TKDE}, 2020.

\bibitem{li2020neural}
J.~Li, A.~Sun, and Y.~Ma, ``Neural named entity boundary detection,''
  \emph{TKDE}, 2020.

\bibitem{sutskever2014sequence}
I.~Sutskever, O.~Vinyals, and Q.~V. Le, ``Sequence to sequence learning with
  neural networks,'' in \emph{NeurIPS}, 2014.

\bibitem{devlin2019bert}
J.~Devlin, M.-W. Chang, K.~Lee, and K.~Toutanova, ``Bert: Pre-training of deep
  bidirectional transformers for language understanding,'' in \emph{NAACL},
  2019.

\bibitem{liu2019roberta}
Y.~Liu, M.~Ott, N.~Goyal, J.~Du, M.~Joshi, D.~Chen, O.~Levy, M.~Lewis,
  L.~Zettlemoyer, and V.~Stoyanov, ``Roberta: A robustly optimized bert
  pretraining approach,'' \emph{arXiv}, 2019.

\bibitem{he2020deberta}
P.~He, X.~Liu, J.~Gao, and W.~Chen, ``Deberta: Decoding-enhanced bert with
  disentangled attention,'' in \emph{ICLR}, 2020.

\bibitem{kasai2021deep}
J.~Kasai, N.~Pappas, H.~Peng, J.~Cross, and N.~Smith, ``Deep encoder, shallow
  decoder: Reevaluating non-autoregressive machine translation,'' in
  \emph{ICLR}, 2021.

\bibitem{li2020sentence}
B.~Li, H.~Zhou, J.~He, M.~Wang, Y.~Yang, and L.~Li, ``On the sentence
  embeddings from bert for semantic textual similarity,'' in \emph{EMNLP},
  2020.

\bibitem{wang2018glue}
A.~Wang, A.~Singh, J.~Michael, F.~Hill, O.~Levy, and S.~Bowman, ``Glue: A
  multi-task benchmark and analysis platform for natural language
  understanding,'' in \emph{EMNLP}, 2018.

\bibitem{radford2018improving}
A.~Radford, K.~Narasimhan, T.~Salimans, I.~Sutskever \emph{et~al.}, ``Improving
  language understanding by generative pre-training,'' 2018.

\bibitem{liu2021pre}
P.~Liu, W.~Yuan, J.~Fu, Z.~Jiang, H.~Hayashi, and G.~Neubig, ``Pre-train,
  prompt, and predict: A systematic survey of prompting methods in natural
  language processing,'' \emph{ACM Computing Surveys}, 2021.

\bibitem{wu2021self}
L.~Wu, H.~Lin, C.~Tan, Z.~Gao, and S.~Z. Li, ``Self-supervised learning on
  graphs: Contrastive, generative, or predictive,'' \emph{TKDE}, 2021.

\bibitem{liu2021self}
X.~Liu, F.~Zhang, Z.~Hou, L.~Mian, Z.~Wang, J.~Zhang, and J.~Tang,
  ``Self-supervised learning: Generative or contrastive,'' \emph{TKDE}, 2021.

\bibitem{liu2022graph}
Y.~Liu, M.~Jin, S.~Pan, C.~Zhou, Y.~Zheng, F.~Xia, and P.~Yu, ``Graph
  self-supervised learning: A survey,'' \emph{TKDE}, 2022.

\bibitem{zhong2023self}
Q.~Zhong, L.~Ding, J.~Liu, B.~Du, and D.~Tao, ``Self-evolution learning for
  discriminative language model pretraining,'' in \emph{Findings of ACL}, 2023.

\bibitem{zhong2023revisiting}
Q.~Zhong, L.~Ding, J.~Liu, X.~Liu, M.~Zhang, B.~Du, and D.~Tao, ``Revisiting
  token dropping strategy in efficient bert pretraining,'' in \emph{ACL}, 2023.

\bibitem{joshi2020spanbert}
M.~Joshi, D.~Chen, Y.~Liu, D.~S. Weld, L.~Zettlemoyer, and O.~Levy, ``Spanbert:
  Improving pre-training by representing and predicting spans,'' \emph{TACL},
  2020.

\bibitem{sun2020ernie}
Y.~Sun, S.~Wang, Y.~Li, S.~Feng, H.~Tian, H.~Wu, and H.~Wang, ``Ernie 2.0: A
  continual pre-training framework for language understanding,'' in
  \emph{AAAI}, 2020.

\bibitem{yang2019xlnet}
Z.~Yang, Z.~Dai, Y.~Yang, J.~Carbonell, R.~R. Salakhutdinov, and Q.~V. Le,
  ``Xlnet: Generalized autoregressive pretraining for language understanding,''
  \emph{NeurIPS}, 2019.

\bibitem{gao2021simcse}
T.~Gao, X.~Yao, and D.~Chen, ``Simcse: Simple contrastive learning of sentence
  embeddings,'' in \emph{EMNLP}, 2021.

\bibitem{yan2021consert}
Y.~Yan, R.~Li, S.~Wang, F.~Zhang, W.~Wu, and W.~Xu, ``Consert: A contrastive
  framework for self-supervised sentence representation transfer,'' in
  \emph{ACL}, 2021.

\bibitem{jiang2022promptbert}
T.~Jiang, S.~Huang, Z.~Zhang, D.~Wang, F.~Zhuang, F.~Wei, H.~Huang, L.~Zhang,
  and Q.~Zhang, ``Promptbert: Improving bert sentence embeddings with
  prompts,'' in \emph{EMNLP}, 2022.

\bibitem{clark2019electra}
K.~Clark, M.-T. Luong, Q.~V. Le, and C.~D. Manning, ``Electra: Pre-training
  text encoders as discriminators rather than generators,'' in \emph{ICLR},
  2019.

\bibitem{panda2021shuffled}
S.~Panda, A.~Agrawal, J.~Ha, and B.~Bloch, ``Shuffled-token detection for
  refining pre-trained roberta,'' in \emph{NAACL: Student Research Workshop},
  2021.

\bibitem{lester2021power}
B.~Lester, R.~Al-Rfou, and N.~Constant, ``The power of scale for
  parameter-efficient prompt tuning,'' in \emph{EMNLP}, 2021.

\bibitem{brown2020language}
T.~Brown, B.~Mann, N.~Ryder, M.~Subbiah, J.~D. Kaplan, P.~Dhariwal,
  A.~Neelakantan, P.~Shyam, G.~Sastry, A.~Askell \emph{et~al.}, ``Language
  models are few-shot learners,'' in \emph{NeurIPS}, 2020.

\bibitem{schick2020few}
T.~Schick and H.~Sch{\"u}tze, ``Few-shot text generation with
  pattern-exploiting training,'' in \emph{EMNLP}, 2021.

\bibitem{schick2021exploiting}
T.~Schick and H.~Schutze, ``Exploiting cloze-questions for few-shot text
  classification and natural language inference,'' in \emph{ACL}, 2021.

\bibitem{morris2020textattack}
J.~Morris, E.~Lifland, J.~Y. Yoo, J.~Grigsby, D.~Jin, and Y.~Qi, ``Textattack:
  A framework for adversarial attacks, data augmentation, and adversarial
  training in nlp,'' in \emph{EMNLP}, 2020.

\bibitem{yamaguchi2021frustratingly}
A.~Yamaguchi, G.~Chrysostomou, K.~Margatina, and N.~Aletras, ``Frustratingly
  simple pretraining alternatives to masked language modeling,'' in
  \emph{EMNLP}, 2021.

\bibitem{alajrami2022does}
A.~Alajrami and N.~Aletras, ``How does the pre-training objective affect what
  large language models learn about linguistic properties?'' in \emph{ACL},
  2022.

\bibitem{nijkamp2021script}
E.~Nijkamp, B.~Pang, Y.~N. Wu, and C.~Xiong, ``Script: Self-critic pretraining
  of transformers,'' in \emph{NAACL}, 2021.

\bibitem{hermann2015teaching}
K.~M. Hermann, T.~Kocisky, E.~Grefenstette, L.~Espeholt, W.~Kay, M.~Suleyman,
  and P.~Blunsom, ``Teaching machines to read and comprehend,'' in
  \emph{NeurIPS}, 2015.

\bibitem{narayan2018don}
S.~Narayan, S.~B. Cohen, and M.~Lapata, ``Don’t give me the details, just the
  summary! topic-aware convolutional neural networks for extreme
  summarization,'' in \emph{EMNLP}, 2018.

\bibitem{liu2021topic}
J.~Liu, Y.~Zou, H.~Zhang, H.~Chen, Z.~Ding, C.~Yuan, and X.~Wang, ``Topic-aware
  contrastive learning for abstractive dialogue summarization,'' in
  \emph{Findings of EMNLP}, 2021.

\bibitem{Ott:19}
M.~Ott, S.~Edunov, A.~Baevski, A.~Fan, S.~Gross, N.~Ng, D.~Grangier, and
  M.~Auli, ``fairseq: {A} fast, extensible toolkit for sequence modeling,'' in
  \emph{NAACL-HLT}, W.~Ammar, A.~Louis, and N.~Mostafazadeh, Eds., 2019.

\bibitem{rothe2020leveraging}
S.~Rothe, S.~Narayan, and A.~Severyn, ``Leveraging pre-trained checkpoints for
  sequence generation tasks,'' \emph{TACL}, 2020.

\bibitem{zhong2022improving}
Q.~Zhong, L.~Ding, L.~Shen, P.~Mi, J.~Liu, B.~Du, and D.~Tao, ``Improving
  sharpness-aware minimization with fisher mask for better generalization on
  language models,'' in \emph{Findings of EMNLP}, 2022.

\bibitem{ng2014conll}
H.~T. Ng, S.~M. Wu, T.~Briscoe, C.~Hadiwinoto, R.~H. Susanto, and C.~Bryant,
  ``The conll-2014 shared task on grammatical error correction,''
  \emph{CoNLL-2014}, 2014.

\bibitem{Chollampatt:18}
S.~Chollampatt and H.~T. Ng, ``A multilayer convolutional encoder-decoder
  neural network for grammatical error correction,'' in \emph{AAAI}, 2018.

\bibitem{Dahlmeier:12}
D.~Dahlmeier and H.~T. Ng, ``Better evaluation for grammatical error
  correction,'' in \emph{NAACL}, 2012.

\bibitem{zhang2018personalizing}
S.~Zhang, E.~Dinan, J.~Urbanek, A.~Szlam, D.~Kiela, and J.~Weston,
  ``Personalizing dialogue agents: I have a dog, do you have pets too?'' in
  \emph{ACL}, 2018.

\bibitem{lidailydialog}
Y.~Li, H.~Su, X.~Shen, W.~Li, Z.~Cao, and S.~Niu, ``Dailydialog: A manually
  labelled multi-turn dialogue dataset,'' in \emph{IJCNLP}, 2017.

\bibitem{alamri2019audio}
H.~Alamri, V.~Cartillier, A.~Das, J.~Wang, A.~Cherian, I.~Essa, D.~Batra, T.~K.
  Marks, C.~Hori, P.~Anderson \emph{et~al.}, ``Audio visual scene-aware
  dialog,'' in \emph{CVPR}, 2019.

\bibitem{rashkin2019towards}
H.~Rashkin, E.~M. Smith, M.~Li, and Y.-L. Boureau, ``Towards empathetic
  open-domain conversation models: A new benchmark and dataset,'' in
  \emph{ACL}, 2019.

\bibitem{papineni2002bleu}
K.~Papineni, S.~Roukos, T.~Ward, and W.-J. Zhu, ``Bleu: a method for automatic
  evaluation of machine translation,'' in \emph{ACL}, 2002.

\bibitem{dong2019unified}
L.~Dong, N.~Yang, W.~Wang, F.~Wei, X.~Liu, Y.~Wang, J.~Gao, M.~Zhou, and H.-W.
  Hon, ``Unified language model pre-training for natural language understanding
  and generation,'' in \emph{NeurIPS}, 2019.

\bibitem{ratner2017snorkel}
A.~Ratner, S.~H. Bach, H.~Ehrenberg, J.~Fries, S.~Wu, and C.~R{\'e}, ``Snorkel:
  Rapid training data creation with weak supervision,'' in \emph{VLDB}, 2017.

\bibitem{liu2019multi}
X.~Liu, P.~He, W.~Chen, and J.~Gao, ``Multi-task deep neural networks for
  natural language understanding,'' in \emph{ACL}, 2019.

\bibitem{zhang2020semantics}
Z.~Zhang, Y.~Wu, H.~Zhao, Z.~Li, S.~Zhang, X.~Zhou, and X.~Zhou,
  ``Semantics-aware bert for language understanding,'' in \emph{AAAI}, 2020.

\bibitem{conneau2020unsupervised}
A.~Conneau, K.~Khandelwal, N.~Goyal, V.~Chaudhary, G.~Wenzek, F.~Guzm{\'a}n,
  {\'E}.~Grave, M.~Ott, L.~Zettlemoyer, and V.~Stoyanov, ``Unsupervised
  cross-lingual representation learning at scale,'' in \emph{ACL}, 2020.

\bibitem{collins2005clause}
M.~Collins, P.~Koehn, and I.~Ku{\v{c}}erov{\'a}, ``Clause restructuring for
  statistical machine translation,'' in \emph{ACL}, 2005.

\bibitem{ding2021progressive}
L.~Ding, L.~Wang, X.~Liu, D.~F. Wong, D.~Tao, and Z.~Tu, ``Progressive
  multi-granularity training for non-autoregressive translation,'' in
  \emph{Findings of the ACL}, 2021.

\bibitem{berg2012empirical}
T.~Berg-Kirkpatrick, D.~Burkett, and D.~Klein, ``An empirical investigation of
  statistical significance in nlp,'' in \emph{EMNLP}, 2012.

\bibitem{see2017get}
A.~See, P.~J. Liu, and C.~D. Manning, ``Get to the point: Summarization with
  pointer-generator networks,'' in \emph{ACL}, 2017.

\bibitem{liu2019text}
Y.~Liu and M.~Lapata, ``Text summarization with pretrained encoders,'' in
  \emph{EMNLP}, 2019.

\bibitem{bao2020unilmv2}
H.~Bao, L.~Dong, F.~Wei, W.~Wang, N.~Yang, X.~Liu, Y.~Wang, J.~Gao, S.~Piao,
  M.~Zhou \emph{et~al.}, ``Unilmv2: Pseudo-masked language models for unified
  language model pre-training,'' in \emph{ICML}, 2020.

\bibitem{krishna2021does}
K.~Krishna, J.~P. Bigham, and Z.~C. Lipton, ``Does pretraining for
  summarization require knowledge transfer?'' in \emph{Findings of EMNLP},
  2021.

\bibitem{shin2020autoprompt}
T.~Shin, Y.~Razeghi, R.~L. Logan~IV, E.~Wallace, and S.~Singh, ``Autoprompt:
  Eliciting knowledge from language models with automatically generated
  prompts,'' in \emph{EMNLP}, 2020.

\bibitem{conneau2018senteval}
A.~Conneau and D.~Kiela, ``Senteval: An evaluation toolkit for universal
  sentence representations,'' in \emph{LREC}, 2018.

\bibitem{hao2019modeling}
J.~Hao, X.~Wang, B.~Yang, L.~Wang, J.~Zhang, and Z.~Tu, ``Modeling recurrence
  for transformer,'' in \emph{NAACL}, 2019.

\end{thebibliography}

\begin{IEEEbiography}[{\includegraphics[width=1in,height=1.25in,clip,keepaspectratio]{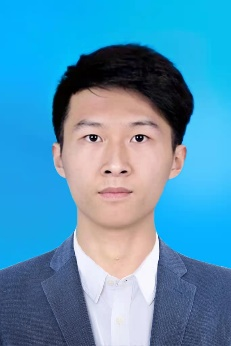}}]{Qihuang Zhong}
	is currently pursuing the Ph.D. degree in Artificial Intelligence from the School of Computer Science, Wuhan University. His research interests include language model pretraining, natural language understanding and generation. He has authored or co-authored over 10 research papers at top conferences and international journals, including ACL, EMNLP, COLING, IEEE TKDE, IEEE/ACM TASLP and \textit{etc}. He won the general language understanding (GLUE) and more difficult language understanding (SuperGLUE) challenges.
\end{IEEEbiography}

\begin{IEEEbiography}[{\includegraphics[width=1in,height=1.25in,clip,keepaspectratio]{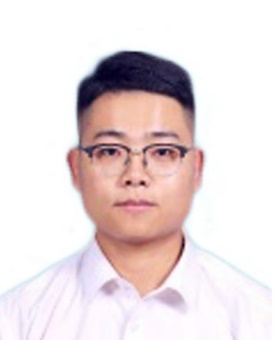}}]{Liang Ding} received Ph.D. from the University of Sydney. He is currently an algorithm scientist with JD.com and leading the NLP research group at JD Explore Academy. He works on deep learning for NLP, including language model pretraining, language understanding, generation, and translation. He published over 50 research papers in NLP/AI, including ACL, EMNLP, ICLR, ICML, AAAI, IJCAI, SIGIR, CVPR, IEEE TPAMI, and TKDE, and importantly, some of his works were successfully applied to the industry. He served as Area Chair and Session Chair for ACL 2022 and SDM 2021. He won many AI challenges, including SuperGLUE/ GLUE, WMT2022, IWSLT 2021, WMT 2020, and WMT 2019. Liang led the team to be the first to outperform human performance (in Dec. 2021) on two challenging tasks and then got first place (in Jan. 2022) with an average score of 91.3 on the general language understanding evaluation (GLUE) benchmark. Afterward, Liang's team developed the Vega-v2 model, which sat atop the SuperGLUE leaderboard (in Oct. 2022).
\end{IEEEbiography}

\begin{IEEEbiography}[{\includegraphics[width=1in,height=1.25in,clip,keepaspectratio]{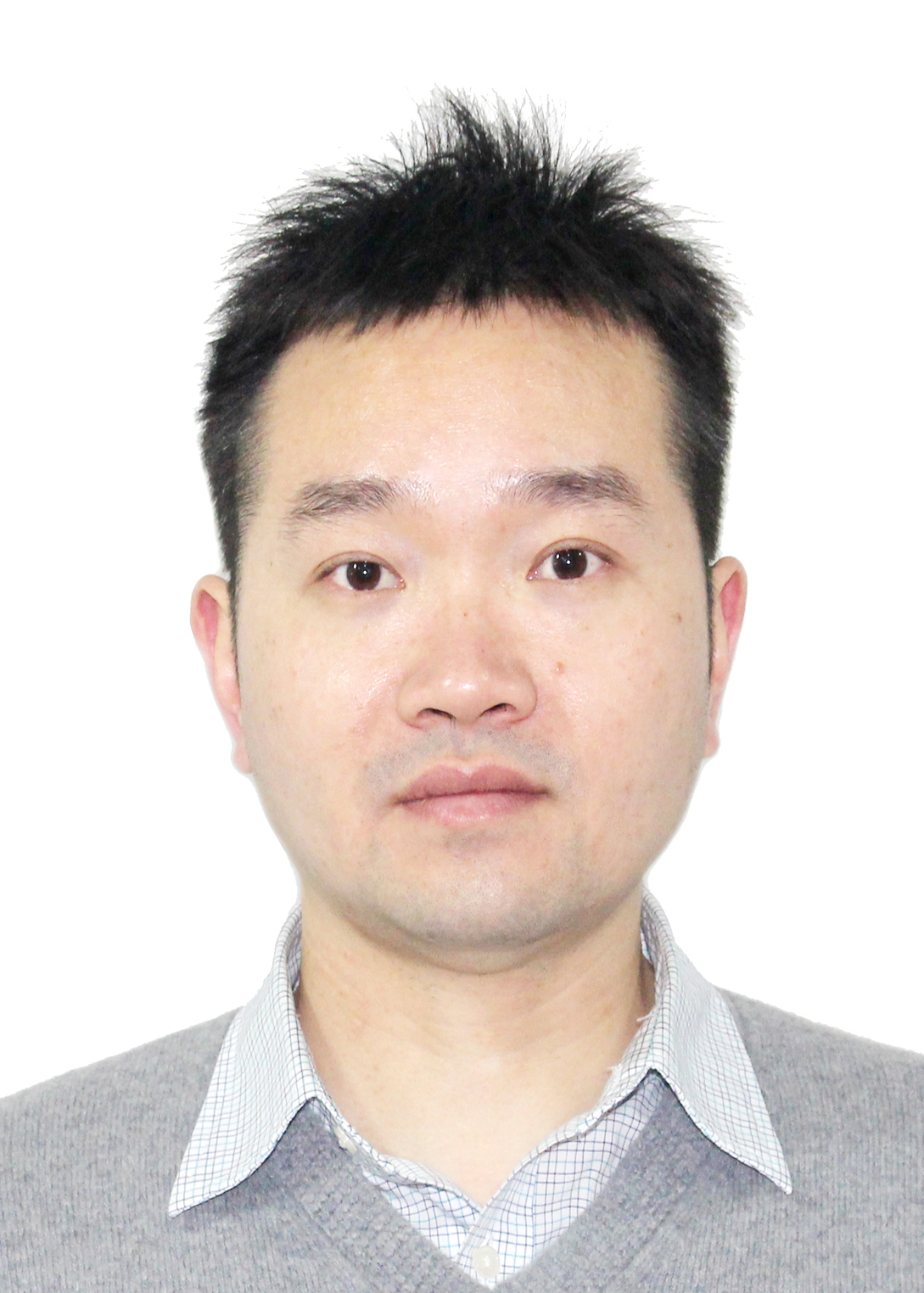}}]{Juhua Liu} is currently a professor with the School of Computer Science and Institute of Artificial Intelligence, Wuhan University. His research interests mainly include image processing, computer vision, natural language processing and machine learning. He has published more than 40 research papers in CV/NLP/AI, including IJCV, IEEE TIP, IEEE TKDE, IEEE/ACM TASLP, CVPR, ACL, AAAI, IJCAI, ACM MM and EMNLP, \textit{etc}. He serves as a reviewer of several top journals, including IEEE TPAMI, IEEE TIP, IEEE TCYB, IEEE TNNLS, IEEE TASLP, IEEE TIM, \textit{etc}, and regularly serves as PC member of CVPR, AAAI, IJCAI, ACM MM, ICASSP and ICME, \textit{etc}. \end{IEEEbiography}

\begin{IEEEbiography}[{\includegraphics[width=1in,height=1.25in,clip,keepaspectratio]{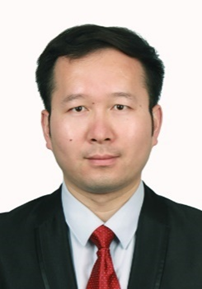}}]{Bo Du} (M'10-SM'15) 
is currently a professor with the School of Computer Science and Institute of Artificial Intelligence, Wuhan University. He is also the director of National Engineering Research Center for Multimedia Software, Wuhan University, Wuhan, China. He has more than 80 research papers published in the IEEE TPAMI, TIP, IEEE TCYB, IEEE TGRS, \textit{etc}. Fourteen of them are ESI hot papers or highly cited papers. His major research interests include machine learning, computer vision, and image processing. He is currently a senior member of IEEE and serves as associate editor for Neural Networks, Pattern Recognition and Neurocomputing. He won the Highly Cited Researcher (2019\textbackslash2020) by the Web of Science Group. He won IEEE Geoscience and Remote Sensing Society 2020 Transactions Prize Paper Award, the IJCAI (International Joint Conferences on Artificial Intelligence) Distinguished Paper Prize, IEEE Data Fusion Contest Champion, and IEEE Workshop on Hyperspectral Image and Signal Processing Best paper Award.
\end{IEEEbiography}

\begin{IEEEbiography}[{\includegraphics[width=1in,height=1.25in,clip,keepaspectratio]{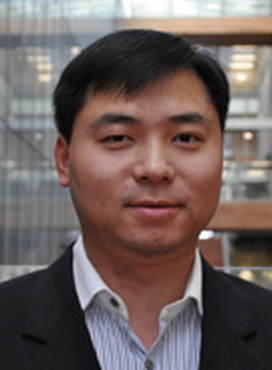}}]{Dacheng Tao} (F'15) is currently a professor of computer science and an ARC Laureate Fellow in the School of Computer Science and the Faculty of Engineering at The University of Sydney. He mainly applies statistics and mathematics to artificial intelligence and data science. His research is detailed in one monograph and over 200 publications in prestigious journals and proceedings at leading conferences. He received the 2015 Australian Scopus-Eureka Prize, the 2018 IEEE ICDM Research Contributions Award, and the 2021 IEEE Computer Society McCluskey Technical Achievement Award. He is a fellow of the Australian Academy of Science, AAAS, ACM, and IEEE.\end{IEEEbiography}

\end{document}